\documentclass[journal]{IEEEtran}

\usepackage{graphicx}
\usepackage{epstopdf}
\usepackage{subcaption}
\usepackage[table,xcdraw]{xcolor}
\usepackage{hyperref}
\usepackage{times}
\usepackage{amsmath}
\usepackage{amssymb}
\usepackage{array}
\usepackage{overpic}
\usepackage{float}
\usepackage[ruled,linesnumbered]{algorithm2e}
\usepackage{enumitem}
\usepackage{booktabs}
\usepackage[english]{babel}
\usepackage{amsthm}
\usepackage{svg}
\usepackage{adjustbox}
\usepackage{multirow}

\graphicspath{ {picture/} }

\begin{document}

\title{Adaptive 3D Gaussian Splatting Video Streaming: Visual Saliency-Aware Tiling and Meta-Learning-Based Bitrate Adaptation}

\author{
Han Gong, Qiyue Li,~\IEEEmembership{Senior Member,~IEEE}, Jie Li,~\IEEEmembership{Member,~IEEE}, Zhi Liu,~\IEEEmembership{Senior Member,~IEEE}

\thanks{Han Gong and Qiyue Li is with the School of Electrical Engineering and Automation, Hefei University of Technology, Hefei, China, and also with the Engineering Technology Research Center of Industrial Automation of Anhui Province, Hefei, China (e-mail: han\_gong@mail.hfut.edu.cn, liqiyue@mail.ustc.edu.cn)}

\thanks{Jie Li is with the School of Computer Science and Information Engineering,
Hefei University of Technology, Hefei, China (e-mail: lijie@hfut.edu.cn).
}

\thanks{Zhi Liu is with The University of Electro-Communications, Tokyo, Japan
(e-mail: liu@ieee.org).
}

}

\maketitle

\begin{abstract}

3D Gaussian splatting video (3DGS) streaming has recently emerged as a research hotspot in both academia and industry, owing to its impressive ability to deliver immersive 3D video experiences. However, research in this area is still in its early stages, and several fundamental challenges—such as tiling, quality assessment, and bitrate adaptation—require further investigation.
In this paper, we tackle these challenges by proposing a comprehensive set of solutions. Specifically, we propose an adaptive 3DGS tiling technique guided by saliency analysis, which integrates both spatial and temporal features. Each tile is encoded into versions possessing dedicated deformation fields and multiple quality levels for adaptive selection. We also introduce a novel quality assessment framework for 3DGS video that jointly evaluates spatial-domain degradation in 3DGS representations during streaming and the quality of the resulting 2D rendered images. Additionally, we develop a meta-learning-based adaptive bitrate algorithm specifically tailored for 3DGS video streaming, achieving optimal performance across varying network conditions.
Extensive experiments demonstrate that our proposed approaches significantly outperform state-of-the-art methods.
The source code will be made publicly available upon acceptance of the paper.

\end{abstract}


\IEEEpeerreviewmaketitle

\section{INTRODUCTION}

With the advancement of multimedia and communication technologies, volumetric video, delivered through VR/AR/MR, has seen rapid development and been widely adopted in multiple fields. For example, in telemedicine, volumetric video provides surgeons with high-quality visual feedback, allowing remote execution of complex and urgent medical procedures while also expanding to auditory and haptic feedback \cite{desselle2020augmented,liu2021point}. In 3D video conferencing, it facilitates true immersive communication, allowing users to interact with one another in virtual environments as they would in the real world \cite{jansen2020pipeline}.  %To better achieve practical application outcomes, the development of volumetric video has focused primarily on two key aspects: higher visual quality and more convenient interactive usability \cite{viola2023volumetric}.        

Volumetric video representations are primarily divided into explicit forms (such as point clouds and meshes) and implicit representations (exemplified by Neural Radiance Fields, NeRF) \cite{jin2023capture}. Explicit representations demonstrate superior editability, facilitating the implementation of various interactive functions for volumetric video. However, both point clouds and meshes exhibit significant visual quality limitations: point clouds lose substantial textural and geometric details due to their discrete nature, while meshes can only represent smooth surfaces and fail to accurately render sharp or complex structures. In comparison, NeRF achieves photorealistic rendering quality, but its implicit radiance field representation imposes severe constraints on editing efficiency, rendering it impractical for real-time applications \cite{yuan2022nerf}. As an innovative volumetric video representation,  3D Gaussian Splatting (3DGS) combines NeRF-level scene realism with the editing flexibility inherent in explicit representations \cite{ye2024gaussian}. By leveraging adaptive 3D Gaussian primitives with anisotropic covariance, this approach achieves real-time rendering through differentiable splatting while maintaining geometric fidelity. The explicit yet parameterized representation allows selective manipulation of scene elements at varying granularities, overcoming the opacity of implicit neural representations. This hybrid paradigm not only preserves high-frequency texture details comparable to NeRF's volumetric rendering, but also enables efficient geometric transformations akin to point cloud operations \cite{jiang2024robust}. For diverse volumetric video applications, 3DGS-based volumetric video undoubtedly represents the most promising research focus currently and its efficient streaming is one fundamental research challenge in these volumetric video applications.

However, conventional volumetric video transmission methods cannot be directly applied to 3DGS videos due to fundamental incompatibilities between the unique data characteristics of 3DGS and existing technical frameworks\cite{sun2025lts}. The unstructured Gaussian distribution in 3DGS disrupts the structured spatial assumptions inherent to traditional point clouds or meshes, while the high-dimensional attributes of each Gaussian primitive far exceed the geometric color dimensions of conventional volumetric videos \cite{wu2024recent}. Additionally, due to the unique rendering mechanism of 3DGS, different Gaussian primitives exhibit varying rendering weights, with those possessing higher rendering weights significantly impacting visual quality \cite{lee2024compact}. Compounded by novel distortion types exemplified through Gaussian overlap artifacts and spherical harmonic discontinuities arising from anisotropic rendering characteristics that cannot be effectively quantified by conventional quality metrics these fundamental discrepancies collectively demand a dedicated compression transmission and quality evaluation framework tailored to 3DGS video \cite{martin2025gs}.

To enable specialized streaming for 3DGS video, researchers have conducted studies and proposed several novel transmission systems \cite{sun20243dgstream} \cite{sun2024multi} \cite{wang2024v}. Nevertheless, these systems lack sufficient consideration of the intrinsic characteristics of 3DGS video, particularly in terms of effective visual feature extraction, video quality assessment, and adaptive bitrate (ABR) algorithm design. Quality assessment for 3DGS must jointly evaluate geometric fidelity and rendered perceptual quality. Existing metrics focus on single representations, failing to capture their combined impact \cite{sun20243dgstream} \cite{sun2025lts}. While recent works have made progress in streaming static 3DGS models through attribute quantization and Level of Detail (LOD) control, these methods prove inadequate for dynamic 3DGS video scenarios \cite{tsai2025l3gs} \cite{chen2025pcgs}. The temporal coherence of Gaussian attributes introduces unique compression dynamics that static optimization frameworks cannot capture. 

%Existing ABR algorithms for 3DGS streaming struggle to generalize across diverse network conditions \cite{sun2025lts}.
%传统体积视频传输方法在套用至3DGS视频时难以解决逐帧数据量过大的问题\cite{sun2025lts}，而动态3DGS的重建方法虽然极大压缩了单帧的数据量，但受限于解码时间会在客户端造成播放的卡顿\cite{sun20243dgstream}。通常的做法会将体积视频构造成两个版本，一个直接传输重建模型来避免过长解码时间，一个传输需要解码的数据来规避过大的单帧数据量。这种传输模式套用在3DGS streaming时会造成实际传输数据量上限和下限差异巨大的问题，因为3DGS视频有着比点云视频为代表的传统体积视频的更大的单帧数据量。这为我们设计能够适用于不同网络环境的3DGS视频ABR算法带来了困难。
Conventional volumetric video transmission methods fail to address the prohibitive per-frame data volumes when applied to 3DGS video \cite{sun2025lts}. Although dynamic 3DGS reconstruction techniques significantly compress frame-level data, decoding latency constraints cause client-side playback stuttering \cite{sun20243dgstream}. The standard dual-mode transmission strategy—delivering either pre-reconstructed models to circumvent decoding delays or encoded data to reduce bandwidth—creates extreme disparities between minimum and maximum transmission demands for 3DGS streaming \cite{li2022optimal}. This implementation gap arises because 3DGS videos exhibit per-frame data volumes substantially larger than traditional volumetric formats exemplified by point cloud videos, fundamentally constraining the design of network-agnostic ABR algorithms for 3DGS streaming systems. Although meta-learning has demonstrated potential for network adaptation in conventional video streaming \cite{li2023metaabr} \cite{bentaleb2024bitrate}, its application to 3DGS remains fundamentally limited by the inherent complexity of Gaussian representations: The dynamic interdependence between geometric attributes and perceptual quality obstructs effective meta-knowledge transfer, while sparse 3DGS training data prevents models from capturing cross-scenario adaptation patterns.

The current challenges in 3DGS video streaming can be summarized as follows:

\begin{itemize}

 \item \textbf{Visual Saliency Extraction in 3DGS Video:} Similar to conventional volumetric video, viewers consistently prefer to focus on the most visually salient content. However, as an emerging video format, 3DGS video presents unique challenges for visual saliency extraction due to its complex attribute composition and distinctive rendering mechanism. Currently, there is a lack of effective methods specifically designed for saliency detection in 3DGS content. Consequently, traditional approaches commonly employed in volumetric video streaming - such as viewpoint prediction and salient region segmentation - prove inadequate when directly applied to 3DGS video streaming scenarios.

 \item \textbf{Quality Assessment Criteria for 3DGS Video:} As a novel video representation technology, 3DGS fundamentally differs from conventional formats in its scene representation paradigm. Unlike point clouds, voxels, or meshes that rely solely on 3D model geometry, or NeRF that depends exclusively on rendered outputs without explicit models, 3DGS employs a hybrid approach. It combines explicit Gaussian primitives with their splatting-based rendered results to achieve comprehensive scene representation. This unique characteristic necessitates a dual evaluation framework that simultaneously considers both the geometric fidelity of the underlying Gaussian model and the perceptual quality of the final rendered output. Consequently, establishing a dedicated quality assessment standard becomes imperative for 3DGS video transmission systems.

%当前的3DGS流媒体技术大多集中于构建更适用于传输的3DGS视频源，尚未对专为3DGS内容设计的ABR算法进行充分研究。LTS虽然提出了一种适用于3DGS的ABR算法\cite{sun2025lts}，但缺乏全面的用户体验评价，且对带宽的需求过高限制了在通用场景下的泛化。此外基于强化学习的ABR算法在适用于3DGS streaming时受限于双版本传输的巨大上下限数据量差异，且在视频源数据量不足时表现欠佳。这严重影响了训练模型在不同网络条件和部署场景下的泛化能力。
 \item \textbf{Adaptive Bitrate Selection for 3DGS Video:} Current 3DGS streaming research predominantly focuses on constructing transmission-optimized video sources, while ABR algorithms specifically designed for 3DGS content remain insufficiently investigated \cite{zhu2025dynamic}. Although LTS proposed a 3DGS-compatible ABR framework \cite{sun2025lts}, it lacks comprehensive user experience evaluation and imposes excessive bandwidth requirements that limit generalization in common scenarios. Furthermore, reinforcement learning-based ABR algorithms face fundamental limitations when applied to 3DGS streaming: the significant variation between minimum and maximum data volumes in dual-version transmission systems creates adaptation barriers, while inadequate video source data severely compromises trained models' generalization capabilities across diverse network conditions and deployment scenarios.

%针对上述的3DGS视频streaming过程中存在的挑战，我们设计了一套3DGS视频streaming系统来解决这些困难，并实现了3DGS视频streaming的实际应用。我们的贡献可以被总结为以下几点：

%1、我们设计了一种充分利用3dgs复杂属性来表达视觉显著性的特征提取方法，并在此基础上设计了基于3dgs显著性的自适应切块技术。通过我们的方法可以精准预测视频中用户的感兴趣区域，并在streaming过程中针对显著性与视角预测结果对这些区域进行高质量推流。
%2、我们设计了一种全新的3dgs视频质量评估方法，该方法综合考虑了streaming过程中3dgs本身在空间域上的质量损失以及渲染后的2D图像质量损失。在该质量评估方法的基础上我们考量了不同高斯点以及不同属性值对于视频质量的影响，并设计了一种基于切块的多质量版本视频构建方法，该方法能实现在不同的带宽场景下推流适合的视频质量版本。
%3、我们设计了一种基于元学习的3dgs视频ABR算法，该方法首先对点云视频在多种通信环境下进行ABR训练，仅需少量3dgs视频流数据便可以迁移得到3dgs视频的ABR模型。结合上述提到的视角预测结果以及多质量视频版本，我们可以实现不同通信环境下的最优3dgs视频streaming。

%1、我们设计了一种3DGS视频的自适应切块方法。构造了一个针对3DGS视频的显著性特征提取网络，通过不同区块的显著性差异我们将均匀的切块自适应聚合为符合用户视觉关注的新切块。

\end{itemize}

To address the aforementioned challenges in 3DGS video streaming, we have designed a comprehensive 3DGS video streaming system that effectively resolves these difficulties and enables practical deployment of 3DGS video streaming applications. Our system fundamentally advances the state-of-the-art by simultaneously addressing three critical aspects of 3DGS streaming: attention-aware content delivery, hybrid quality evaluation, and network-adaptive transmission. The principal contributions of our work can be summarized as follows:
\begin{itemize}

\item We propose an adaptive tiling method for 3DGS video. A saliency feature extraction network tailored for 3DGS content is designed to estimate the visual importance of different regions. Based on saliency variations across uniform tiles, we adaptively merge them into new tiles that better align with users’ visual attention.

\item We develop a tile-based dynamic 3DGS encoding method compatible with adaptive tiling. By analyzing motion patterns across GoFs, tiles are categorized into static, low-dynamic, and high-dynamic types, enabling efficient deformation-based reconstruction. In parallel, we introduce a saliency-aware quality tiering strategy that applies Gaussian pruning with variable rates, ensuring high visual quality under constrained bandwidth.

\item We introduce a meta-reinforcement learning-based ABR scheme for 3DGS video, supported by a hierarchical perceptual Quality of Experience (QoE) model that reflects holistic perceptual quality. This approach enables robust adaptation across diverse network conditions while efficiently optimizing user experience.

\item We conduct extensive simulations across diverse network conditions and 3DGS video
datasets, and the results verify the performance of the proposed schemes.

\end{itemize}

The rest of the paper is organized as follows: Section 2 critically analyzes related work in 3DGS volumetric video streaming, and adaptive bitrate algorithms. Section 3 presents our end-to-end streaming framework, integrating three innovations: saliency-driven adaptive tiling, tile-based dynamic 3DGS encoding with multi-quality tiering, and meta-learning-based QoE optimization. Each segmentation phase within our framework is meticulously detailed in Section 4. Finally, Section 5 validates the framework through extensive experiments across diverse network conditions and 3DGS video datasets.

\section{RELATED WORK}

This section describes related work on 3DGS content streaming and ABR schemes for volumetric video.

\subsection{3DGS Content Streaming}
%由于3DGS内容的高视觉质量和易于编辑的特性，使得3DGS内容的高效streaming成为了一个重点研究内容。对于3DGS内容的streaming主要分为两个方向，对于静态3DGS内容的streaming和对于动态内容的streaming。

%在对静态3DGS内容的streaming研究中，L3GS定制化训练流程，通过迭代剪枝与分层冻结策略生成可控规模的3DGS模型，对场景分割为语义对象并构建基础层加增强层的分层结构，实现了渐进式传输与视口自适应调度。PCGS提出渐进式压缩框架，通过渐进掩码策略逐步解码新锚点，并采用三平面量化细化现有锚点属性，结合上下文建模提升熵编码效率。其联合优化锚点数量与质量的设计，在压缩性能上媲美单速率方法（如HAC++），同时支持动态网络带宽下的渐进增强。StreamGS通过逐帧预测像素对齐的3D高斯，并利用相邻帧的可靠匹配信息，动态聚合高斯集合，实现了无姿态图像流的在线通用化重建，摆脱对SfM预处理或深度先验的依赖。

%动态3DGS内容由于更广泛的应用前景受到了更多研究者的关注。在目前已有的动态3DGS内容传输框架中，3DGStream[1]通过神经变换缓存（NTC）建模3D高斯的运动属性，结合动态高斯自适应生成策略处理场景新增物体，实现了动态自由视点视频的在线实时构建。Dynamic 3DGS Streaming[2]提出了一种动态3DGS流媒体的多帧比特率分配方法，通过建模几何、球谐系数、不透明度及变换属性的率失真特性，设计了基于梯度上升的模型驱动算法（MGA）及其自适应版本（MGAA），实现在动态网络带宽约束下最大化渲染质量。V3[3]提出将动态3D高斯属性编码为2D视频流，利用硬件视频编解码器实现高效压缩与移动端实时渲染。其创新点包括时空一致性保持的训练策略、残差熵损失和时序损失，有效降低了存储需求。TGH[4]提出了一种名为时间高斯层次结构（Temporal Gaussian Hierarchy, TGH）的新方法，用于高效建模长时间体视频。该方法通过分析动态场景中不同区域的时间冗余性，构建多层次的4D高斯基元，自适应地共享静态或缓慢变化的区域信息，从而显著减少存储需求和计算成本。IGS[5]提出锚点驱动高斯运动网络（AGM-Net），通过多视图光流特征投影和锚点邻域插值解码运动，单次推断即可生成帧间高斯变形，结合关键帧优化与最大点数约束策略有效抑制了误差累积。EvolvingGS[6]通过动态演化的高斯表示与高效压缩策略，显著提升了动态场景重建的质量与效率，通过局部调整与全局优化相结合，平衡时间连贯性与细节还原能力。

The high visual quality and inherent editability of 3DGS content have made its efficient streaming a critical research focus. Current 3DGS streaming approaches primarily bifurcate into two directions: static 3DGS content streaming and dynamic content streaming.

In static 3DGS streaming research, L3GS \cite{tsai2025l3gs} introduces a customized training pipeline that generates controllable-scale 3DGS models through iterative pruning and hierarchical freezing strategies. By segmenting scenes into semantic objects and constructing a base layer plus enhancement layer architecture, it achieves progressive transmission and viewport-adaptive scheduling. PCGS \cite{chen2025pcgs} proposes a progressive compression framework that incrementally decodes new anchors via progressive masking while refining existing anchor attributes through tri-plane quantization. Enhanced by context modeling, this approach boosts entropy coding efficiency. Its joint optimization of anchor quantity and quality achieves compression performance comparable to single-rate methods like HAC++, while supporting progressive enhancement under dynamic network bandwidth. StreamGS \cite{li2025streamgs} enables online generalized reconstruction from unposed image streams by predicting pixel-aligned 3D Gaussians frame-by-frame and dynamically aggregating Gaussian sets using reliable inter-frame matching information, eliminating dependencies on SfM preprocessing or depth priors.

Dynamic 3DGS content has attracted significant research attention due to its broader application prospects \cite{zhu2025dynamic}. Among existing dynamic 3DGS transmission frameworks, 3DGStream \cite{sun20243dgstream} employs a Neural Transformation Cache (NTC) to model motion attributes of 3D Gaussians, combining dynamic Gaussian adaptive generation strategies to handle newly added objects in scenes, achieving real-time online construction of dynamic free-viewpoint videos. Dynamic 3DGS Streaming \cite{sun2024multi} proposes a multi-frame bitrate allocation method for dynamic 3DGS streaming, designing a model-driven algorithm (MGA) and its adaptive variant (MGAA) through rate-distortion optimization of geometric, spherical harmonic, opacity, and transformation attributes. This approach maximizes rendering quality under dynamic network bandwidth constraints. V3 \cite{wang2024v} encodes dynamic 3D Gaussian attributes into 2D video streams, leveraging hardware video codecs for efficient compression and real-time mobile rendering. Its innovations include a temporally consistent training strategy, residual entropy loss, and temporal loss, significantly reducing storage requirements. TGH \cite{xu2024representing} introduces a Temporal Gaussian Hierarchy (TGH) for efficient long-form volumetric video representation. By analyzing temporal redundancy across dynamic scene regions, it constructs hierarchical 4D Gaussian primitives to hierarchically share static or slowly varying regions, substantially reducing storage and computational costs. IGS \cite{yan2025instant} proposes an Anchor-Driven Gaussian Motion Network (AGM-Net), decoding inter-frame Gaussian deformations via multi-view optical flow feature projection and anchor neighborhood interpolation. With key frame optimization and maximum point count constraints, it effectively suppresses error accumulation, enabling single-inference frame interpolation. EvolvingGS \cite{zhang2025evolvinggs} enhances dynamic scene reconstruction quality through evolving Gaussian representations and efficient compression strategies. By integrating local adjustments with global optimization, it balances temporal coherence and detail preservation, achieving significant improvements in dynamic 3DGS encoding efficiency.

%虽然目前针对3DGS视频流已经有部分相关工作，但是绝大部分工作依然集中于如何构建更适用于Streaming的3DGS视频源上。而对于传统体积视频流中较为重要的视频源处理以及优化传输却少有涉及。本文将目光集中于如何更好地传输已有的3DGS视频源上，与现有的各类方案形成了很好的互补，并为3DGS视频流的实际应用提供了新的参考。
While recent research has yielded preliminary investigations into 3DGS video streaming, the overwhelming majority of efforts remain concentrated on reconstructing streamable 3DGS video sources. Critical aspects essential to traditional volumetric video streaming—specifically, source processing and optimized transmission—have received scant attention. This work shifts focus toward efficiently transmitting existing 3DGS video content, thereby establishing complementary advances to current methodologies and offering novel pathways for practical implementation of 3DGS video streaming systems.

\begin{figure*}[htb]
    \centering
    \includegraphics[width=0.7\textwidth]{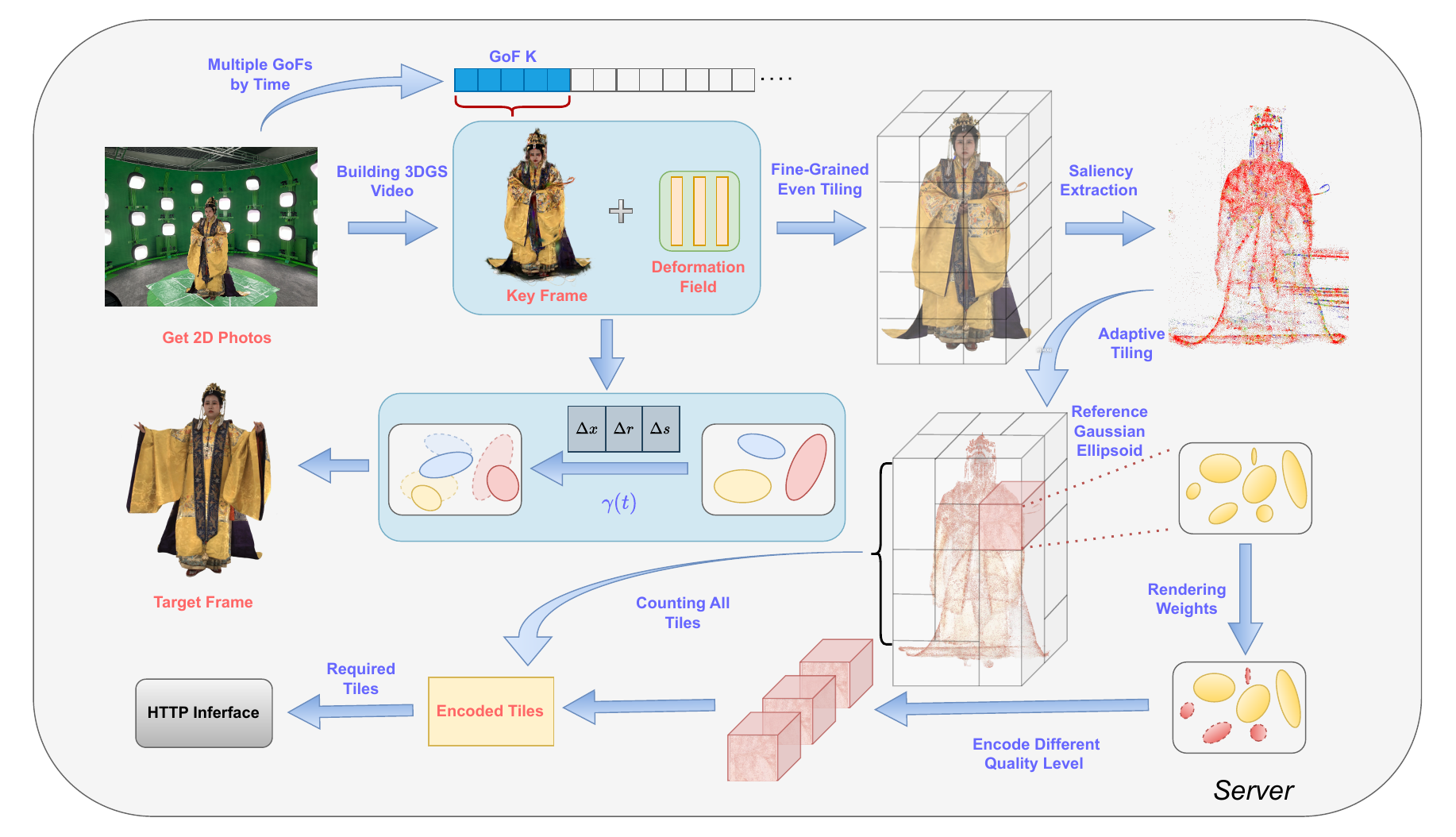}
    \caption{Server-side architecture for 3DGS video streaming.}
    \label{fig_Server}
\end{figure*}

\subsection{ABR Schemes for Volumetric Video}

%体积视频的ABR算法由2D视频的ABR算法扩展而来，与传统2D视频不同的是体积视频往往拥有大得多的单帧数据量，且最终6 DOF的观看需求使得其优化算法更加复杂。对于2D视频而言选择不同的分辨率并且获取对应的视觉质量是一件容易的事，但是在体积视频中，往往需要更复杂的质量区分方法，例如通过切块或者分层的手段去构造不同的视频质量，在衡量视频的传输损失时也需要考虑模型在几何结构等方面的变化。且在考虑环境限制时，体积视频也需要考虑除了带宽条件以外更多的信息，例如用户观看视频时的FoV信息取决于体积视频的哪一部分会被优先传输。类似V-PCC或者G-PCC的编码方法由于其过长的解码时间也是体积视频ABR算法中难以忽视的问题之一。

ABR algorithms for volumetric video inherit principles from 2D video streaming but face amplified challenges due to massive data volumes, 6-degree-of-freedom (6DoF) viewing constraints, and hybrid quality evaluation requirements. Traditional ABR algorithms of 2D video relies on discrete resolution-quality pairs, whereas volumetric video streaming demands sophisticated adaptation strategies. 

Point cloud video, as the most exemplary volumetric video format, has received extensive research attention, with ABR algorithms constituting a major research focus within this domain. In particular, Wang et al. \cite{wang2022qoe} proposed a novel perspective-projection-based QoE model that holistically integrates view frustum, distance, occlusion, and screen resolution, and designs a greedy-based rate adaptation algorithm that transforms the optimization problem into a submodular maximization task, achieving near-optimal QoE with low complexity for tile-based point cloud video streaming. Li et al. \cite{li2023toward} studied rolling prediction-optimization-transmission framework that leverages short-window prediction to mitigate long-term bandwidth/FoV prediction errors and employs a serial DRL solver SC-DDQN for real-time ABR decisions, significantly improving QoE in point cloud video streaming. Zhang et al. \cite{zhang2021efficient} designed a QoE-driven joint network-computation adaptation framework that dynamically adjusts per-patch download quality and super-resolution ratios to balance bandwidth consumption, computational load, and visual consistency in volumetric video streaming. Huang et al. \cite{huang2022toward} proposed an AI-native DRL-based adaptive streaming framework that dynamically selects lightweight encoder-decoder models to balance network conditions, device computation, and QoE. Liu et al. \cite{liu2023cav3} proposed a client-cache-assisted viewport adaptive streaming framework that leverages long/mid/short-term viewport trajectory prediction to prioritize caching of temporally repetitive tiles. A progressive frame patching scheme with KKT-optimized tile rate allocation for FoV-adaptive point cloud video streaming was proposed by Zong et al. \cite{zong2025progressive}. Shi et al. \cite{shi2024qv4} studied a QoE-based viewpoint-aware tiling and adaptive bitrate allocation scheme for V-PCC-encoded volumetric video streaming.

Although 3DGS employs an explicit spatial representation analogous to point clouds, its underlying data structure and rendering mechanisms are fundamentally distinct. Consequently, quality assessment methodologies based on geometric loss metrics prove inadequate for evaluating the perceptual experience of 3DGS video. To address this critical disconnect, we develop a dedicated ABR algorithm tailored for 3DGS video streaming, establishing a viewer-centric QoE standard that accurately reflects authentic user viewing experiences.

%对于新兴的neural volumetric video，同样有研究者做了ABR算法上的尝试。V²NeRF \cite{shi2024towards} 将NeRF的隐式表示特性与传统ABR机制深度融合，针对NeRF生成与点云投影争抢GPU资源而造成的计算资源与网络带宽的耦合竞争问题，提出了两阶段解耦ABR算法，分别优化不同资源维度。并利用NeRF驱动的ABR决策优化,通过NeRF特性降低ABR复杂度，实现了全场景体积视频的高效streaming。LTS \cite{sun2025lts} 针对3DGS视频的特性提出了一个全新的ABR框架，通过将3DGS帧分解为基础层加增强层，基础层保留场景几何结构，增强层添加高频细节来适配3DGS的多层结构,扩展静态LapisGS \cite{shi2024lapisgs} 至动态场景。同时采用了分tile传输以及分段传输来适配3DGS视频的空间局部性与帧间相关性。通过联合优化层-块-段三维决策实现了ABR算法在3DGS视频streaming上的实际应用。
Emerging research has explored ABR algorithms for neural volumetric video. For example, V²NeRF \cite{shi2024towards} unified the implicit representation properties of NeRF with conventional ABR mechanisms to address computational resource contention between NeRF rendering and point cloud projection on GPUs. The study proposed a two-stage decoupled ABR algorithm that independently optimized distinct resource dimensions. LTS \cite{sun2025lts} introduced a novel ABR framework specifically designed for 3DGS video, extending the static LapisGS framework \cite{shi2024lapisgs} to dynamic scenarios. Through joint optimization of layer-tile-segment decision dimensions, the implementation of ABR algorithms for practical 3DGS video streaming was achieved.

%虽然LTS非常优秀地将传统体积视频的ABR方法结合进了3DGS streaming当中，但是依然缺乏对用户观看习惯以及3DGS本身视觉属性的考量。并且所需传输的单帧数据量过大限制了其在实际场景下的应用，对带宽的需求非常苛刻。相较之下我们的方案更多地考虑了用户的实际观看体验，并且追求在通常网络环境下的普适性。
While researchers had integrated conventional volumetric video ABR methods into 3DGS streaming systems, these approaches failed to account for user viewing behavior patterns and the inherent visual characteristics of 3DGS representations. Furthermore, the prohibitively large per-frame data volumes required for transmission imposed stringent bandwidth requirements that severely limited practical deployment in real-world scenarios. In contrast, our solution prioritizes authentic viewing experience while ensuring broader applicability under common network conditions.

\section{OVERVIEW} 

%本文以3DGS视频流为研究对象设计了一种全新的视频流传输框架。有别于之前的3DGS视频streaming工作将重心集中于如何构造更适用于streaming的3DGS视频源,我们将目光集中在如何更高效地对已有的视频源进行传输上。Figure \ref{fig_Server}中展示了我们的Server-side architecture for 3DGS video streaming，我们的框架包含了从视频源处理到最终优化传输的所有步骤，为3DGS视频streaming的实际应用提供了参考。具体而言我们的工作可以分为三个主要组成部分：
%我们将视频源按照时间切分为多个GoF，每个GoF内我们通过显著性提取将视频处理为符合users’ visual attention的自适应切块。对于每个切块我们按照其动态程度分别使用高斯变形场对其进行编码，并按照切块在渲染和显著性上的重要程度对每个切块进行multi-quality tiering。为了更好地将传统体积视频的ABR算法适配于3DGS视频我们构建了一个专门的QoE标准用于评价用户在观看3DGS视频时的实际体验，并且结合元强化学习来加强我们算法在不同网络条件下的普适性。通过挑选能最大化QoE的切块及其对应质量版本，我们实现了用户在实际观影中的最佳体验。
This study proposes a novel streaming framework specifically designed for 3DGS video. Unlike prior 3DGS streaming research primarily focused on reconstructing streamable 3DGS video sources, our work centers on optimizing the transmission efficiency of existing 3DGS video content. Figure \ref{fig_Server} illustrates the server-side architecture of our 3DGS video streaming system, encompassing all processing stages from source preparation to optimized delivery, thereby providing a practical foundation for real-world 3DGS video streaming applications. The system architecture is composed of three major functional stages: (1) Saliency-driven Adaptive Tiling, (2) Tile-based Dynamic 3DGS Encoding with Multi-quality Tiering, and (3) Meta-learning-based QoE Optimization. 

We temporally partition the video source into multiple Groups of Frames (GoFs). Within each GoF, visual saliency extraction guides the processing of video content into adaptive tiles aligned with users' visual attention patterns. For every tile, Gaussian deformation fields are selectively applied according to its motion-intensity class determined by displacement magnitude thresholds, while multi-quality tiering is implemented based on rendering significance and saliency importance. To bridge conventional volumetric video ABR algorithms with 3DGS streaming, we establish a dedicated QoE metric that quantifies authentic viewing experiences during 3DGS video playback. This framework integrates meta-reinforcement learning to enhance cross-environment generalizability. By strategically selecting tiles and their corresponding quality levels to maximize QoE, we achieve optimal viewing experiences under real-world conditions.
The following section elaborates in detail on each component of our proposed 3DGS video streaming framework.

\begin{figure*}[htb]
    \centering
    \includegraphics[width=0.8\textwidth]{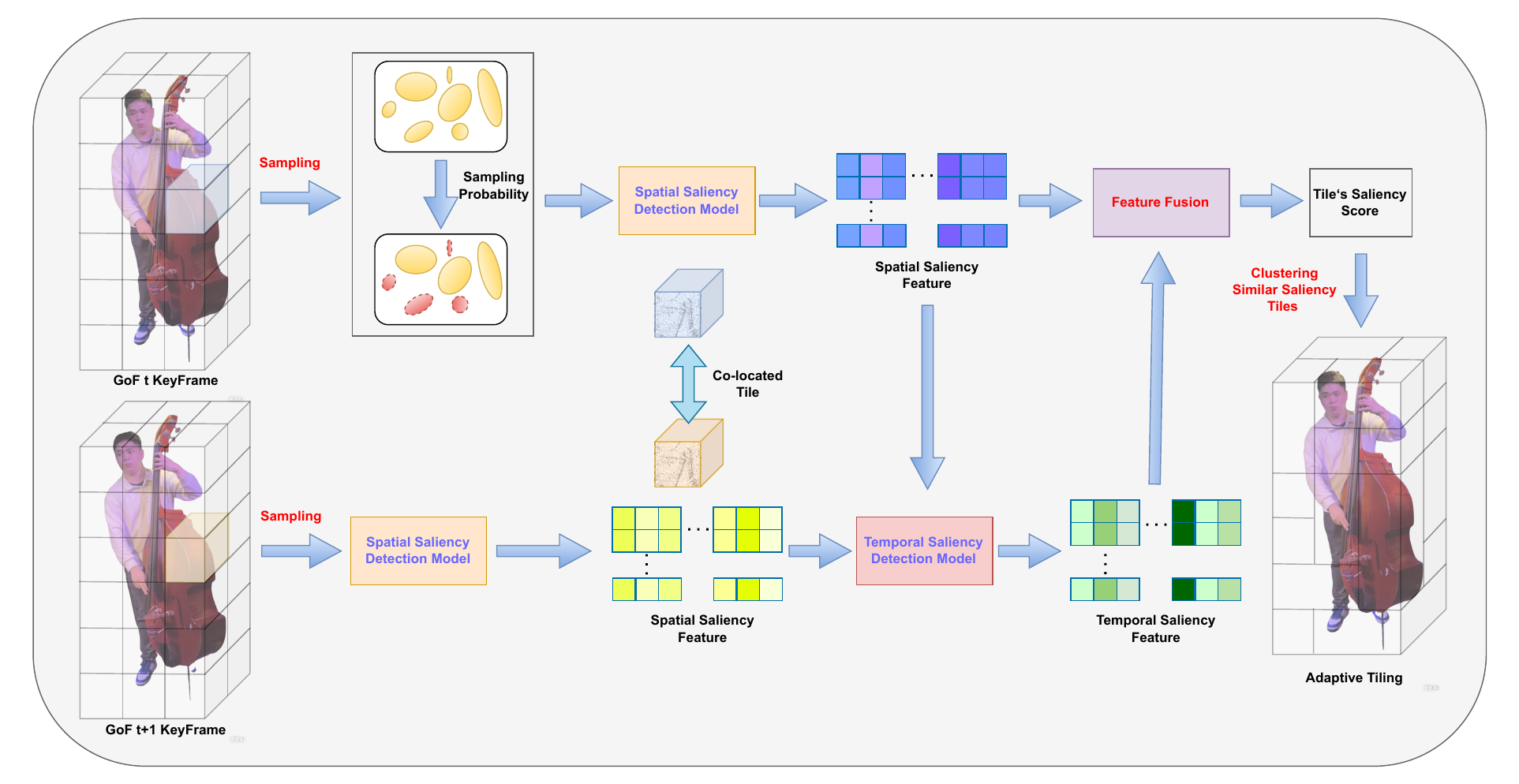}
    \caption{Saliency extraction and adaptive tiling.}
    \label{fig_tiling}
\end{figure*}

\section{DESIGN OF SYSTEM}
\subsection{Saliency-driven Adaptive Tiling}
%体积视频通常由于过大的数据量导致在传输时不得不只选取其中一部分进行传输，通常的做法会是通过切块来将模型区分为多个独立的小范围。在传输时我们按照视频的显著性，或者是用户观看的FoV范围去选取最合适的切块进行传输，对于3DGS视频而言切块的方法同样适用。然而均匀切块通常会在切块数量上面临抉择，过少的切块会导致FoV的匹配程度降低，导致传输范围过大超出用户观看范围。过多的分块则会加重3DGS视频在解码上的资源开销。因而基于视频内容和用户行为动态调整分块形状可以最大化内容与用户关注区域的匹配，同时平衡计算与通信资源。In this paper, we propose a saliency-driven adaptive tiling method.方法的具体流程如图\ref{fig_tiling}所示。
Volumetric video streaming often requires transmitting only a subset of data due to its massive volume. A common approach is to partition the model into smaller independent tiles. During transmission, tiles are selected based on the user’s FoV. This tiling strategy also applies to 3DGS videos. However, uniform tiling faces a trade-off in tile count: too few tiles reduce FoV alignment accuracy, leading to excessive transmission beyond the user’s viewing range, while too many tiles increase decoding overhead and encoding inefficiency. Dynamically adjusting tile shapes based on video content and user behavior can maximize content-FoV alignment while balancing computational and communication resources. In this paper, we propose a saliency-driven adaptive tiling method, with the detailed workflow illustrated in Figure \ref{fig_tiling}.

%我们首先将初始的3DGS模型按照帧组均匀地切分为多个小切块，我们用$t_{j, t}$表示帧组（GoF）t中的第j个切块。然后在每个切块$t_{j, t}$内我们基于渲染权重对高斯基元进行采样。我们将采样概率定义为：
This method first partitions the initial 3DGS model into small tiles per GoF, denoted as $t_{j, t}$ for the $j$-th tile in GoF $t$. Within each tile $t_{j, t}$, Gaussian primitives are sampled based on rendering weights. The sampling probability $p_{\text {sample }}(i)$ is defined as:
\begin{equation}
    p_{\text {sample }}(i)=\frac{w_i}{\sum_{n=1}^N w_n},
\end{equation}
\begin{equation}
    w_i=\sigma_i \cdot \sqrt{\operatorname{det}\left(\Sigma_i\right)}.
\end{equation}
%其中$w_i$表示第i个高斯基元在整个切块中的渲染权重，描述其在整个切块中的重要程度。$\sigma_i$为$t_{j, t}$中第i个高斯基元的不透明度，描述其在渲染中的可见性。$\Sigma_i$为第i个高斯基元的协方差矩阵，描述其在三维空间中的形状和方向。我们通过$\sqrt{\operatorname{det}\left(\Sigma_i\right)}$去作为高斯基元的等效体积。通过量化每个高斯基元对渲染结果的视觉重要性，我们在采样过程中优先保留体积大且不透明度高的高斯基元，采样后的切块我们定义为$D_{j, t}$。对于每个$D_{j, t}$我们分别设计了spatial saliency detection model和temporal saliency detection model去提取其空间和时间上的显著性。
The parameter $w_i$ represents the rendering weight of the $i$-th Gaussian primitive within tile $t_{j, t}$, quantifying its relative importance for the tile's visual synthesis; $\sigma_i$ denotes the opacity of this primitive, governing its visibility contribution during rendering; while $\Sigma_i$ constitutes the covariance matrix characterizing the primitive's 3D spatial configuration through shape anisotropy and orientation. The term $\sqrt{\operatorname{det}(\Sigma_i)}$ approximates the effective volume of the primitive. We define the visual importance of each primitive as $w_i = \sigma_i \sqrt{\operatorname{det}(\Sigma_i)}$, which jointly considers both opacity and spatial extent. Based on this importance metric, we prioritize sampling primitives with higher $w_i$ values. The resulting sampled tile is denoted as $D_{j,t}$. For each $D_{j,t}$, we further design a spatial saliency detection model and a temporal saliency detection model to extract spatial and temporal saliency cues, respectively.

%由于每个高斯基元都是通过高阶SH系数去生成对应视角下的颜色信息的，但是对于特征提取而言我们难以通过视角信息去规定一个高斯基元的颜色，因为在观看过程中用户的视角是实时变化的，因此我们采用每个高斯基元的零阶SH系数$c_{i, R}^0, c_{i, G}^0, c_{i, B}^0$去近似高斯基元的颜色信息。因此对于$D_{j, t}$中的每个高斯基元i的位置属性和颜色属性可以分别表示为：$p_{i, t}=\left\{x_{i, t}, y_{i, t}, z_{i, t}\right\}$和$a_{i, t}=\left\{c_{i, t, R}^0, c_{i, t, G}^0, c_{i, t, B}^0\right\}$。我们首先通过全连接层（FC layer）提取点i的初始特征:
\textbf{Spatial Saliency Detection Model.} Each Gaussian primitive generates view-dependent color through high-order spherical harmonics (SH) coefficients. However, due to the continuously changing user FoVs during viewing, it is impractical to determine a primitive's color from a specific FoV for feature extraction. Therefore, we approximate the color of each Gaussian primitive using its zero-order SH coefficients. Specifically, $c_{i, t, R}^0$, $c_{i, t, G}^0$, and $c_{i, t, B}^0$ represent the zero-order SH coefficients corresponding to the red, green, and blue color channels, respectively, and are used as an approximation of the primitive’s RGB color. Thus, the position and color attributes of the $i$-th primitive in $D_{j, t}$ are:
\begin{equation}
    p_{i, t}=\left\{x_{i, t}, y_{i, t}, z_{i, t}\right\},
\end{equation}
\begin{equation}
    a_{i, t}=\left\{c_{i, t, R}^0, \\ c_{i, t, G}^0, c_{i, t, B}^0\right\} .
\end{equation}
Initial features for primitive $i$ are extracted via a fully connected (FC) layer:
\begin{equation}
    f_{i, t}=F C\left(p_{i, t} \oplus a_{i, t}\right),
\end{equation}
where $\oplus$ denotes concatenation.
%这样我们可以得到$D_{j, t}$的初始特征$f_{j, t}$，然后通过我们设计的local discrepancy catcher (LDC)模块来统一整合空间和颜色差异。具体而言我们首先使用Neighborhood Coding unit显式编码$D_{j, t}$中每个高斯基元i与其邻域高斯基元k在坐标和颜色上的差异，生成增强的高斯基元特征。其中邻域高斯基元k的位置和颜色属性为$p_{i, t}^k$和$a_{i, t}^k$。我们首先将零阶SH系数转换为灰度值来反映人类视觉对颜色的敏感度：
The Local Discrepancy Catcher (LDC) module integrates spatial and color discrepancies. Specifically, the neighborhood coding unit explicitly encodes coordinate and color differences between Gaussian primitive $i$ and its neighboring primitive $k$ in $D_{j,t}$. Grayscale conversion is applied to zero-order SH coefficients to align with human visual sensitivity:
\begin{equation}
    d_{i, t}=0.299 \cdot c_{i, t, R}^0+0.587 \cdot c_{i, t, G}^0+0.114 \cdot c_{i, t, B}^0.
\end{equation}
%则坐标与颜色差异可以显示编码为：
The discrepancy between primitives $i$ and $k$ is then encoded as:
\begin{equation}
    \begin{aligned}
d p_{i, t}^k= & M L P\left[p_{i, t} \oplus p_{i, t}^k \oplus\left(p_{i, t}-p_{i, t}^k\right) \oplus\left\|p_{i, t}-p_{i, t}^k\right\|\right. \\
& \left.\oplus d_{i, t} \oplus d_{i, t}^k \oplus\left(d_{i, t}-d_{i, t}^k\right) \oplus\left\|d_{i, t}-d_{i, t}^k\right\|\right]
\end{aligned},
\end{equation}
%将编码后的差异特征$d p_{i, t}^k$与邻域点的初始特征$f_{i, t}^k$拼接，生成增强特征$\hat{f_{i, t}^{k}}$：
where $p_{i, t}^k$ and $d_{i, t}^k$ are the position and grayscale values of neighbor $k$, respectively. The encoded discrepancy $d p_{i, t}^k$ is concatenated with the neighbor’s initial feature $f_{i, t}^k$, generating an enhanced feature $\hat{f_{i, t}^{k}}$.
\begin{equation}
    \hat{f_{i, t}^{k}}=d p_{i, t}^k \oplus f_{i, t}^k.
\end{equation}
%对于高斯基元i我们将其所有K个相邻基元进行编码并获得最终的增强特征集合为$A_{i, t}=\left\{\hat{f_{i, t}^1}, \hat{f_{i, t}^2}, \hat{f_{i, t}^3} \ldots \hat{f_{i, t}^k}, . ., \hat{f_{i, t}^K}\right\}$。对于每个邻域基元的增强特征$\hat{f_{i, t}^{k}}$我们使用共享MLP生成原始注意力分数$S_{i, t}^k$：
For Gaussian primitive $i$, all $K$ neighboring primitives are encoded, forming an enhanced feature set:
\begin{equation}
    A_{i, t}=\left\{\hat{f_{i, t}^1}, \hat{f_{i, t}^2}, \ldots, \hat{f_{i, t}^K}\right\}.
\end{equation}
A shared Multilayer Perceptron (MLP) generates raw attention scores for each enhanced neighbor feature $\hat{f_{i, t}^k}$:
\begin{equation}
    S_{i, t}^k=\gamma\left(\hat{f_{i, t}^k}, W\right),
\end{equation}
%其中权重$W$在所有邻域高斯基元间共享，$\gamma()$为shared MLP function。通过Softmax函数归一化注意力分数来生成权重$\sigma\left(S_{i, t}^k\right)$：
where $W$ is shared across neighbors, and $\gamma()$ denotes the shared MLP function. Scores are normalized via Softmax:
\begin{equation}
    \sigma\left(S_{i, t}^k\right)=\frac{\exp \left(S_{i, t}^k\right)}{\sum_{k=1}^K \exp \left(S_{i, t}^k\right)}).
\end{equation}
%加权求和邻域特征，得到聚合后的特征$\hat{f_{i, t}}$:
Neighboring features are aggregated via weighted summation:
\begin{equation}
    \hat{f_{i, t}}=\sum_{i=1}^K\left[\hat{f_{i, t}^k} * \sigma\left(S_{i, t}^k\right)\right].
\end{equation}
%通过注意力机制动态调整邻域特征的贡献，使网络更关注显著性较高的区域。

%此外我们还设计了dilated residual block去扩大每个高斯基元的感受野，从而捕捉更大范围的局部上下文信息。通过对输入特征依次执行两次邻域编码和注意力池化操作（Attention Pooling），使每个高斯基元能够接收最多$K^2$个邻域基元的信息，其核心操作可以表示为：
To expand the receptive field, dilated residual blocks iteratively apply neighborhood coding and attention pooling twice, enabling each primitive to capture contextual information from up to $K^2$ neighbors. This process is formalized as:
\begin{equation}
    \left.\hat{F_{j, t}^c}=R_c\left\{LDC_c\left[\hat{F_{j, t}^{c-1}}\right], \theta_u\right]\right\},
\end{equation}
\begin{equation}
    \hat{F_{j, t}^0}=L_1\left(F_{j, t}\right),
\end{equation}
%其中$LDC_c$表示第c层的LDC模块，包含邻域编码、注意力池化等操作，$R_c$表示第c层的降采样操作，$\hat{F_{j, t}^{c-1}}$表示第$c-1$层$D_{j, t}$的输入特征，$\hat{F_{j, t}^c}$为第c层dilated residual block的输出特征，$\theta_u$则为可学习权重。
%where $\hat{F_{j, t}^c}$ is the input feature from the $c-1$-th layer and $\hat{F_{j, t}^c}$ is the output feature from the $c$-th dilated residual block, $R_c$ denotes downsampling, and $\theta_u$ represents learnable weights. Finally, encoded features are decoded through MLP, upsampling, and FC layers to produce spatial saliency feature $FS_t$ for each Gaussian primitive. $t_{j, t}$.
where $\hat{F_{j, t}^c}$ is the input feature from the $(c-1)$-th layer and $\hat{F_{j, t}^c}$ is the output feature from the $c$-th dilated residual block, $R_c$ denotes downsampling, and $\theta_u$ represents learnable weights. Finally, encoded features are decoded through MLP, upsampling, and FC layers to produce spatial saliency feature $F_S^{(t)}$.
%%%
%编码后的特征通过MLP层、上采样层和全连接层进行解码，最终输出$t_{j, t}$的空间显著性特征$FS_{j,t}$。

%\textbf{Temporal saliency detection model.} 对于时间显著性，我们通过temporal contrast layer（TC）与LDC的组合来进行获取。首先我们使用TC提取相邻GoF同位置切块的特征差异来进行表示切块的空间显著性，特征差异反映了从$D_{j, t-1}$到$D_{j, t}$的变化程度。首先我们对$D_{j, t-1}$与$D_{j, t}$提取其global features，具体操作可以表示为：
\textbf{Temporal Saliency Detection Model.} For temporal saliency detection, we combine the temporal contrast layer (TC) with the LDC module to capture dynamic changes between consecutive GoFs. The TC layer extracts feature differences between co-located tiles in adjacent GoFs, reflecting the degree of variation from $D_{j, t-1}$ to $D_{j, t}$. The process begins by extracting global feature $Q_t^c$ from corresponding tiles using max pooling:
\begin{equation}
    Q_t^c=M\left(T_t^c\right),
\end{equation}
\begin{equation}
    Q_{t-1}^c=M\left(T_{t-1}^c\right),
\end{equation}
%其中$T C_{t-1}^c$和$T C_t^c$表示$D_{j, t-1}$与$D_{j, t}$在$c$-th TC的输入，其初始值分别为$\hat{F_{j, t-1}^{1}}$和$\hat{F_{j, t}^{1}}$。我们通过最大池化操作$M()$去获取$T C_{t-1}^c$和$T C_t^c$的全局特征$Q_{t-1}^c$以及$Q_t^c$。对应的全局特征通过其相似性分数$S_{sim}$来计算显著性强度$O_s$：
where $T_{t-1}^c$ and $T_t^c$ represent inputs to the $c$-th TC layer for $D_{j, t-1}$ and $D_{j, t}$, initialized as $\hat{F_{j, t-1}^{1}}$ and $\hat{F_{j, t}^{1}}$ outputs from the first LDC layer, and $M()$ denotes the max pooling operation. The similarity between these global features is computed via a shared MLP:
\begin{equation}
    S_{sim}=\gamma\left(Q_t^c \oplus Q_{t-1}^c\right),
\end{equation}
and converted to a saliency intensity score $O_s$:
\begin{equation}
    O_s=\frac{1}{1+\exp (S_{sim})}+1.
\end{equation}
%我们输入特征按照显著性强度$O_s$进行加权计算，并通过LDC处理后传输至下一层TC模块：
This score dynamically weights the input features of the current frame, amplifying regions with significant temporal changes:
\begin{equation}
    \hat{T_t^c}=O_s * T_t^c.
\end{equation}
The weighted features $\hat{T C_t^{c+1}}$ are then processed through the LDC module and downsampled via $R_c$:
\begin{equation}
    \left.\hat{T_t^{c+1}}=R_c\left\{LDC_c\left[\hat{T_t^c}\right], \theta_u\right]\right\}.
\end{equation}
%在经过多层TC处理后，与前一个GoF相比动态更显著的区域的特征会被逐步放大，因而我们可以有效地捕获对应切块间的时间信息。最后与spatial saliency detection model类似，我们采用MLP, upsampling, and FC layers进行解码，最终输出逐高斯基元的时间显著性特征FT_{t}，$t_{j, t}$的时间显著性特征$FT_{j,t}$
%Through iterative TC layers, dynamically salient regions are progressively highlighted. Finally, the encoded temporal features are decoded using MLP, upsampling, and FC layers, generating temporal saliency features $FT_{j, t}$ for tile $t_{j, t}$.
Through iterative TC layers, dynamically salient regions are progressively highlighted. Finally, the encoded temporal features are decoded using MLP, upsampling, and FC layers, generating temporal saliency feature $F_T^{(t)}$.

%\textbf{Feature Fusion and Adaptive Tiling.} 通过上述两个特征提取模块我们可以获得GoF $t$下的spatial saliency feature $F_S^{(t)}$与temporal saliency features $F_T^{(t)}$。我们使用共享MLP然后通过Softmax函数来计算注意力得分：
\textbf{Feature Fusion and Adaptive Tiling.} Through the aforementioned feature extraction modules, we obtain the spatial saliency feature $F_S^{(t)}$ and temporal saliency feature $F_T^{(t)}$ for GoF $t$. A shared MLP followed by Softmax activation computes attention scores $A_S^{(t)}$ and $A_T^{(t)}$:
%A_S和A_T分别表示空间和时间特征的attention得分
\begin{equation}
    A_S^{(t)}=\sigma\left[\gamma\left(F_S^{(t)}, W_1\right)\right],
\end{equation}
\begin{equation}
    A_T^{(t)}=\sigma\left[\gamma\left(F_T^{(t)}, W_2\right)\right],
\end{equation}
%其中$\sigma()$为Softmax函数，$\gamma()$表示共享MLP操作，$W_1$和$W_2$为可学习权重。
where $\sigma()$ denotes the Softmax function, $\gamma()$ represents the shared MLP operation, and $W_1, W_2$ are learnable weight matrices.

%通过注意力得分我们将空间特征与时间特征进行加权融合并得到新的综合显著性特征（comprehensive saliency feature）$F_C^{t}$：
The comprehensive saliency feature $F_C^{(t)}$ is derived via attention-weighted fusion:
\begin{equation}
    F_C^{(t)}=A_S^{(t)} \odot F_S^{(t)}+A_T^{(t)} \odot F_T^{(t)},
\end{equation}
%其中$\odot$表示逐元素相乘。由此我们可以将切块$t_{j, t}$的综合显著性特征表示为：
with $\odot$ indicating element-wise multiplication. The comprehensive saliency feature for tile $t_{j, t}$ is computed as:
\begin{equation}
    \bar{F}_{C, j}^{(t)}=\frac{1}{N_j} \sum_{i \in \mathcal{B}_j} F_{C, i}^{(t)},
\end{equation}
%其中$\mathcal{B}_j$为切块$t_{j, t}$中的高斯基元的集合，$N_j$为该切块中的基元数量。
where $\mathcal{B}_j$ denotes the set of Gaussian primitives within tile $t_{j, t}$ and $N_j$ is the primitive count.

%我们使用两层全连接网络将聚合特征映射为显著性得分：
%其中$W_a$和$W_b$为可学习权重向量，$b_1$和$b_2$为可学习偏置。真实的显著性得分$\operatorname{Score}_j^(t)$是我们通过统计每个切块在用户观看3DGS视频时的FoV包含频率以及计算inner-tile dispersion和exterior-tile rarity综合标定的\cite{li2022optimal}。
The aggregated comprehensive saliency feature $\bar{F}_{C, j}^{(t)}$ is mapped to a saliency score via a two-layer perceptron:
\begin{equation}
    {S}_j^{(t)} = W_b \left[ \operatorname{ReLU}\left( W_a \bar{F}_{C, j}^{(t)} + b_a \right) \right] + b_b
\end{equation}
where $W_a$, $W_b$ are learnable weights, and $b_a, b_b$ are biases.

We optimize the model using a Smooth $L1$ loss between the predicted score ${S}_j^{(t)}$ and the ground truth saliency score $\operatorname{Score}_j^{(t)}$:
\begin{equation}
    \mathcal{L} = \frac{1}{M} \sum_{j=1}^{M} \ell\left( {S}_j^{(t)} - \operatorname{Score}_j^{(t)}\right)
\end{equation}
\begin{equation}
    \ell(x) = 
    \begin{cases} 
    0.5x^2 & \text{if } |x| < 1 \\
    |x| - 0.5 & \text{otherwise}
    \end{cases}
\end{equation}
The ground truth $\operatorname{Score}_j^{(t)}$ is computed from static saliency detection and dynamic motion estimation as in \cite{li2022optimal}.

%获得所有$t_{j, t}$的显著得分后，我们通过聚类算法将拥有相似显著得分的切块重新聚类到一起，同时拥有高显著值的切块且距离较近的切块会优先被聚类成更大的切块。我们先将每个切块单独归为一类，然后计算每个类之间显著性的相似度，相似度最高的两类被合并为一个新的类。通过不断聚合直至剩余切块数量达到设定阈值时停止合并。
After obtaining the saliency scores for all tiles $t_{j, t}$, we employ a clustering algorithm \cite{pasupathi2021trend} to regroup tiles with similar saliency values. High-saliency tiles in spatial proximity are prioritized for aggregation into larger tiles. The clustering process initiates by treating each tile as an individual cluster. Subsequently, the similarity between every pair of clusters is computed, and the two clusters exhibiting the highest similarity are merged into a new cluster. This iterative merging continues until the number of remaining clusters reaches a predetermined threshold, resulting in adaptively aggregated tiles optimized for saliency coherence and spatial continuity.

\begin{figure*}[htb]
    \centering
    \includegraphics[width=0.7\textwidth]{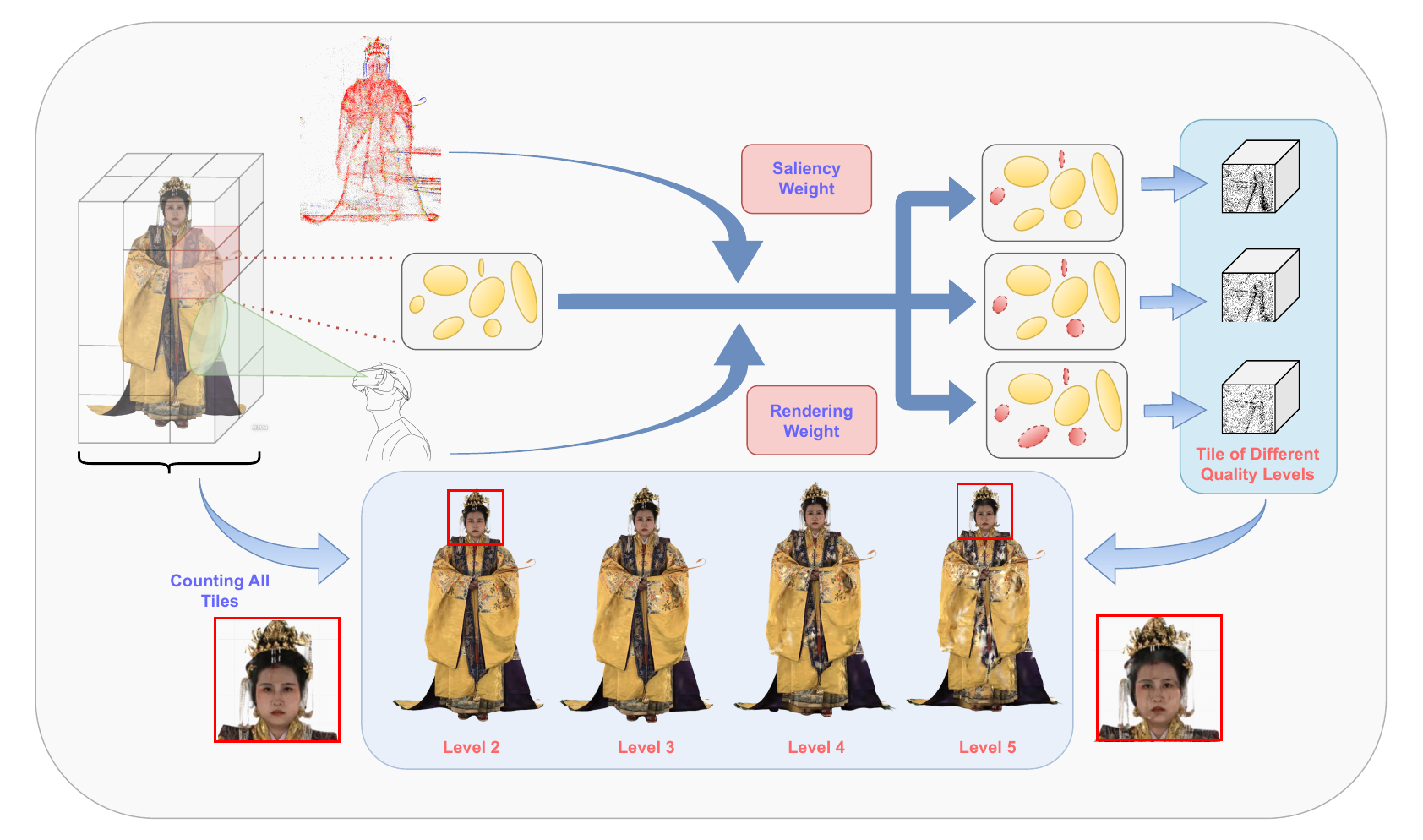}
    \caption{Quality level partitioning.}
    \label{fig_partitioning}
\end{figure*}

\subsection{Tile-based Dynamic 3DGS Encoding and Multi-quality Tiering}
%由于3DGS的显示表达，目前的点云切块技术是可以套用在逐帧的3DGS视频上的。但是3DGS单帧数据量过大的问题限制了逐帧切块传输的普适性，使得3DGS视频streaming对于网络状况的要求极为严苛。对于3DGS视频而言，使用dynamic 3DGS技术引入时间维度，使高斯基元的属性（位置、形状、颜色等）随时间变化，以建模动态场景的运动或形变是一种常见的做法。但是目前所有的dynamic 3DGS方法都是针对整个3DGS模型来进行的，不适用于分切块进行streaming的方式。为了应对这一挑战我们设计了一种动态3DGS切块编码的方法。
Existing point cloud tiling techniques can be applied to frame-wise 3DGS video streaming due to its explicit representation. However, the prohibitively large per-frame data volume severely limits widespread adoption, imposing stringent network requirements \cite{sun2025lts}. For 3DGS video, dynamic 3DGS techniques commonly model temporal evolution by varying Gaussian attributes (position, shape, color) over time to capture scene motion or deformation. Yet all current dynamic 3DGS methods operate on entire 3DGS models and are incompatible with tiled streaming \cite{sun20243dgstream}. To address this challenge, we design a dynamic 3DGS tile encoding method.

%在上一章节中我们通过特征提取网络获得了$t_{j, t}$的显著性得分（saliency score）${S}_j^{(t)}$，对于聚合后的自适应切块$t_{m, t}^a$我们使用其所有聚合前切块的${S}_j^{(t)}$均值${S}_m^{(t)}$作为聚合后显著性得分。我们对比前后两个GoF的所有切块的显著性得分，在前一个GoF里的每个切块$t_{m, t}^a$都在后一个GoF里选择得分最接近的切块$t_{m, t+1}^a$作为其匹配切块。按照匹配切块间的运动向量的模我们将所有切块划分为高动态切块、低动态切块与静止切块三类，这样做的目的是为了避免在编码切块时需要独立对每个切块构造对应的高斯变形场。对于静止切块我们默认其在整个GoF上是没有变化的，可以被视作视频的背景部分，因此不为其配置变形场。对于低动态切块我们将其视作同一物体表面不同部分的协同运动，基于局部运动一致性我们对所有相似切块使用共享变形场而不进行单独配置。高动态切块则独立使用单个变形场对其中所有的切块进行编码，保留复杂运动细节。
Following the acquisition of saliency scores $S_j^{(t)}$ for each tile $t_{j, t}$ via the feature extraction network, aggregated adaptive tiles $t_{m, t}^a$ derive their saliency scores $S_m^{(t)}$ by averaging $S_j^{(t)}$ values across their constituent pre-aggregation tiles. For temporal coherence analysis, each aggregated tile $t_{m, t}^a$ within GoF $t$ is matched to its corresponding tile $t_{m, t+1}^a$ in GoF $t+1$ by selecting the tile with the closest saliency score. Displacement vectors $\mathbf{d}_m$ between matched tile pairs are computed, and tiles are categorized into three motion classes based on the magnitude $\left\|\mathbf{d}_m\right\|$: Static tiles exhibit no motion across GoFs and are treated as background components requiring no deformation fields; low-dynamic tiles demonstrate coordinated surface motions and undergo shared deformation field encoding leveraging local motion consistency; high-dynamic tiles preserve complex independent motions and utilize dedicated deformation fields for optimal reconstruction.

%对低动态切块集合$\mathcal{T}_{\text {low }}$，我们基于特征相似性构建共享组$S_k \subseteq \mathcal{T}_{\text {low }}$，相似性我们采用其余弦相似性$s(m, n)$进行度量：
For low-dynamic tiles $\mathcal{T}_{\text {low }}$, spatially adjacent tiles with similar motion patterns are grouped into shared sets $S_k \subseteq \mathcal{T}_{\text {low }}$ using cosine similarity:
\begin{equation}
    s(m, n)=\frac{\left\langle\mathbf{v}_m, \mathbf{v}_n\right\rangle}{\left\|\mathbf{v}_m\right\|\left\|\mathbf{v}_n\right\|},
\end{equation}
\begin{equation}
    \mathbf{v}_m=\nabla {S}_m^{(t)},
\end{equation}
%$s(m, n)$相近且位置相邻的切块我们划分为同一个低动态切块组。

%对于变形场我们参照 \cite{sun20243dgstream}的方法进行构建，对于切块$t_{m, t}^a$中所有高斯基元的位置$\mu_i \in t_{m, t}^a$我们通过多分辨率哈希网格生成特征：
Deformation fields are constructed following 3DGStreaming's methodology \cite{sun20243dgstream}, where multi-resolution hash grid features for Gaussian positions $\mu_i \in t_{m, t}^a$ are generated as:
\begin{equation}
    h\left(\mu_i\right)=\operatorname{Concat}\left(h\left(\mu_i ; 0\right), h\left(\mu_i ; 1\right), \ldots, h\left(\mu_i ; L-1\right)\right),
\end{equation}
%其中$L$为哈希网格的分辨率层级数，$h\left(\mu_i ; l\right)$为高斯基元位置$\mu_i$在第$l$层哈希网格的编码特征。
with $L$ denoting hash resolution levels and $h\left(\mu_i ; \ell\right)$ representing level-specific encoded features. This hierarchical encoding preserves spatial dependencies while maintaining computational efficiency.

%同样我们使用浅层MLP来输出变换参数：
A shallow MLP generates transformation parameters for Gaussian primitives:
\begin{equation}
    d \mu_i, d q_i=\operatorname{MLP}_m\left(h\left(\mu_i\right)\right),
\end{equation}
%其中$d \mu_i$和$d q_i$分别为MLP预测的高斯基元位置与旋转变化。对于低动态切块组我们共享同一MLP来减少参数，高动态切块我们则使用独立MLP来进行全参数优化。新的高斯基元的位置以及旋转可以表示为：
where $d \mu_i$ and $d q_i$ denote predicted positional and rotational variations. Parameter reduction is achieved through shared MLPs for low-dynamic tile groups, while high-dynamic tiles undergo independent full-parameter optimization via dedicated MLPs. The updated Gaussian primitive positions and rotations are computed as:
\begin{equation}
    \mu_i^{\prime}=\mu_i+d \mu_i,
\end{equation}
\begin{equation}
    q_i^{\prime}=\operatorname{norm}\left(q_i\right) \times \operatorname{norm}\left(d q_i\right).
\end{equation}
%这里的$\operatorname{norm}$表示quaternion normalization，$\times$表示四元数乘法。
Here, $\operatorname{norm}$ denotes quaternion normalization and $\times$ represents quaternion multiplication.

%我们将每个GoF内的所有帧划分为关键帧和目标帧，关键帧通常被设置为每个GoF的第一帧，在传输时我们只针对每个关键帧的切块以及对应的变形场进行编码并传输。目标帧则在传输后再客户端按照变形场进行重构。
We partition all frames within each GoF into key frames and target frames. Key frames are designated as the first frame of each GoF, with only their encoded tiles and corresponding deformation fields transmitted. Target frames are subsequently reconstructed client-side using these deformation fields.

%除此之外，我们仍需对每个切块设置多个不同的质量版本来满足不同通信场景的实际需求。在点云视频的切块传输中，构建一个增量式的分层点云是一种常规做法，这样就可以在传输时选择仅传输基础层抑或同时传输基础层与不同等级的加强层来适应带宽的变化。这种做法同样适用于3DGS视频，有所区别的是3DGS视频由于其渲染模式的特殊性，在确定每个层级需要保留哪些高斯基元时需要更加慎重。我们的切块质量分级方案如 Figure \ref{fig_partitioning}所示。
To accommodate diverse network conditions, multi-quality versions are generated for each tile. Our tile quality partitioning scheme is illustrated in Figure \ref{fig_partitioning}.

%相较于高透明度、大椭球尺寸的高斯点，具有较低透明度与较小椭球尺寸的高斯点对整体渲染质量影响更小。因此我们的核心目标在于识别并过滤这些次要高斯点来构造不同的质量版本。此外不同切块的显著性以及用户在观看视频时可能的视角范围也都是辅助我们进行分层的重要因素。
Primitives with low opacity and small ellipsoidal dimensions contribute minimally to rendering quality compared to high-opacity, large-scale Gaussians. Our core objective is to identify and filter these non-essential primitives across quality levels. Tile saliency further inform hierarchical construction.

%首先对于切块中每个等级的采样概率p_{\text {sample }}(i)，我们仍然采用上一章中的采样概率来进行定义：
The sampling probability $p_{\text {sample }}(i)$ for each primitive in a tile follows our established definition:
\begin{equation}
    p_{\text {sample }}(i)=\frac{w_i}{\sum_{n=1}^N w_n},
\end{equation}
\begin{equation}
    w_i=\sigma_i \cdot \sqrt{\operatorname{det}\left(\Sigma_i\right)}.
\end{equation}

%我们按照不同切块的显著性权重以来优化切块的修剪率。因为用户通常会更注意高视觉显著性的区域，因此我们可以通过降低高显著性切块的修剪率同时加大低显著性切块的修剪率来实现更有利于用户观看的质量分层。我们将所有的显著性得分${S}_m^{(t)}$进行归一化：
Pruning rates are optimized using tile saliency weights. High-saliency regions receive reduced pruning to align with viewer attention patterns, while low-saliency areas undergo more aggressive compression. Normalization is applied as follows:
\begin{equation}
    \tilde{S}_m^{(t)}=\frac{{S}_m^{(t)}-S_{\min }}{S_{\max }-S_{\min }}.
\end{equation}
%对于每个切块的调整后修剪率我们将其计算为：
The adjusted pruning rate per tile is computed as:
\begin{equation}
    p_{\mathrm{adj}}=p_{\mathrm{base}} \cdot\left(1-\tilde{S}_m^{(t)} \cdot \alpha\right),
\end{equation}
%其中p_{\mathrm{base}}为基础修剪率，$\alpha \in[0,1]$为修剪系数用于调节最低修剪率。我们设定所有切块的最低修剪率为8%，这样在最低的质量等级下，整体修剪率依然低于30%，保证了高视觉显著性切块的最低质量等级依然有较好的观看效果。由此我们可以反推出每个切块应该使用的修剪系数$\alpha$。
%经过我们的实验，通过$p_{\text {sample }}(i)$修剪约30%的高斯基元时仍然能够保持较高的视觉质量，而当修剪超过50%时会出现较为严重的视觉质量突变。因此我们对于每层的基础修剪率定为上一层的15%，保证用户在前三个质量等级下依然有好的视觉体验。而最差的质量等级修剪率也不会低于50%，避免了画面质量的突然降低。
where $p_{\mathrm{base}}$ denotes the base pruning rate and $\alpha \in[0,1]$ regulates minimum pruning. Empirical results show that pruning 30\% of Gaussians via $p_{\text {sample }}(i)$ preserves high visual quality \cite{wang2024v}, whereas exceeding 50\% causes severe degradation \cite{tsai2025l3gs}. We therefore implement a base pruning rate of 15\% per layer as $p_{\mathrm{base}}$, ensuring satisfactory quality for the top three levels. The lowest quality level maintains at most 50\% pruning to prevent abrupt quality collapse. We enforce a minimum pruning rate of 8\% across all tiles, ensuring overall pruning remains below 30\% at the lowest quality level. This guarantees acceptable visual fidelity even for high-saliency tiles at minimal quality settings. The pruning coefficient $\alpha$ is derived inversely to satisfy this constraint.

\subsection{Meta-learning-based QoE Maximization Scheme}

%在本节中，我们提出了基于元学习的3DGS视频优化传输模块。我们首先提供3DGS视频的QoE指标的量化描述，然后将ABR问题表示为深度强化学习任务，可以使用元强化学习框架来解决。由于高质量3DGS视频采集依赖昂贵设备，需要多目相机阵列进行拍摄，且需专业人员对FoV轨迹、显著性等信息进行标注，导致数据集获取成本高且规模有限。且目前大部分3DGS视频数据集都并没有按照体积视频的6DoF要求进行拍摄，很多都存在场景覆盖不全，例如只有正面180度可见，而背面并没有进行拍摄。或者场景复杂度单一，例如场景中只存在单个人物且动作幅度受限。这些问题共同导致了3DGS视频数据集在streaming任务上的局限性。而元学习拥有小样本快速适应的优点，通过元学习在多个相关任务上预训练模型，使其在新任务中仅需少量样本即可调整策略，缓解数据不足问题。且通过元学习的知识迁移能力我们可以提升对新场景的泛化能力，在面对视频内容区别较大的新数据集时，依然能够实现较好的决策效果。

In this section, we propose a meta-learning-based optimization module for 3DGS video streaming. We first provide a quantitative description of the QoE metric for 3DGS video and formulate the ABR problem as a deep reinforcement learning task solvable through meta-reinforcement learning frameworks. 

%首先我们需要重新对3DGS视频的用户体验进行定义，现有的体积视频QoE公式不能很好地表现3DGS视频的用户体验，原因在于3DGS视频渲染方式的特殊性，对于点云来说几何质量可以在一定程度上代表视频的最终质量，因为在执行渲染时点云是依照自身的rgb属性来获取渲染图像的，但是对于神经渲染视频我们往往更关注最终渲染画面的质量，在3dgs视频中最终的渲染图像是多个参数共同作用的结果（例如球谐函数，SH系数），这意味着我们在关注几何上的质量损失的同时也需要关注最终画面的渲染质量。针对现有QoE模型在3DGS视频评估中的局限性，我们提出分层感知QoE模型（Layered Perceptual QoE Model),对于GoFt QoE计算公式可以写成：

To address the limitations of existing QoE models in evaluating 3DGS video, we redefine the user experience assessment framework. Unlike conventional point cloud video where geometric fidelity predominantly determines perceived quality—as rendered images derive directly from RGB attributes, 3DGS video rendering quality arises from the combined parameters of spherical harmonics, Gaussian positions and other attributes, meaning that both geometric loss and rendering loss must be jointly considered.

To address the limitations of existing QoE models in evaluating 3DGS video, we propose a layered perceptual QoE model. For GoF $t$, the QoE is computed as:
\begin{equation}
    QoE=\lambda\left(\alpha Q_t^{\mathrm{geo}}+(1-\alpha) Q_t^{\mathrm{render}}\right)-\mu \cdot P_t^d-\sigma \cdot P_t^f,
\end{equation}
where $Q_t^{\text {geo }}$ represents geometric quality, $Q_t^{\text {render }}$ denotes rendering quality, $P_t^d$ indicates stall time, and $P_t^f$ specifies stall frequency. $\lambda$, $\mu$ and $\sigma$ are nonnegative weighting parameters corresponding to the average video quality, stall time and stall frequency, respectively. $\alpha$ is weighting factor determining the contribution ratio of geometric quality to rendering quality.
%其中Q_t^{\mathrm{geo}}为几何质量，Q_t^{\mathrm {render}}为渲染质量，P_t^d为stall time，P_t^f为stall frequency。

%对于几何质量Q_{\mathrm{geo}}的定义为
The geometric quality $Q_t^{\text {geo }}$is computed as:
\begin{equation}
    Q_t^{\text{geo}}=\sum_{k=1}^{K_t} \sum_{r=1}^R\left[\mathrm{PSNR}_{t, k, r} \cdot \Phi_{t,k}\right] \cdot x_{t, k, r},
\end{equation}
where $K_t$ represents the total number of tiles in the current GoF $t$. The term $\mathrm{PSNR}_{t, k, r}$ denotes the Peak Signal-to-Noise Ratio (PSNR) between the rendered tile $k$ at quality level $r$ and its reference version \cite{schwarz2018emerging}, while $\Phi_{t, k}$ represents the spatial influence factor of tile $k$ in GoF $t$, quantifying the spatial importance weight of each tile in 3DGS video. Its core objective is to dynamically adjust the contribution weights of different tiles in QoE computation by integrating visual salience and viewpoint visibility, thereby accurately reflecting how spatial positions in 3DGS rendering impact perceptual quality. This mechanism ensures that tiles with higher visibility or stronger visual prominence are assigned greater weights in the QoE calculation, while occluded or peripheral tiles are downweighted. The $\Phi_{t, k}$ is calculated as:

%\Phi_{t,k}为GoF\t中切块\k的空间影响因子，用于量化3DGS视频中每个切块对最终用户体验质量的空间重要性权重。其核心目标是通过结合视觉显著性和视角可见性，动态调整不同切块在QoE计算中的贡献权重，从而更精准地反映3DGS视频渲染中空间位置对感知质量的影响。x_{t, k, r}为切块质量等级选择变量，其值在0和1之间进行优化，当其为1时表示选择GoF\t中切块\k的第\r个质量等级版本。K_t为GoF\t的切块总数。

%对于空间影响因子\Phi_{t,k}我们定义为：
\begin{equation}
    \Phi_{t,k}=\frac{1}{1+e^{-\gamma \cdot s_{t, k}}} \cdot v_{t, k}.
\end{equation}
%s_{t, k}为切块的显著性权重，\gamma为显著性缩放因子，v_{t, k}为切块i在当前视角下的可见性比例（由视椎体与切块的交集比例得到）。
Here, $s_{t, k}$ reflects the salience weight of tile $k$, determined by its visual prominence in the viewport, and $v_{t, k}$ represents its visibility ratio, computed as the intersection area between the viewport and tile $k$ normalized by the tile’s total area. The parameter $\gamma$ scales the salience weight, and $x_{t, k, r}$ is a binary selection variable constrained by:
\begin{equation}
    \sum_{r=1}^R x_{t, k, r}=1, x_{t, k, r} \in[0,1],
\end{equation}
where $x_{t, k, r}=1$ indicates the selection of the $r$ quality level for tile $k$.

%对于渲染质量Q_{\mathrm {render}}，我们将其定义为：
By predicting the user’s viewport in advance, we can obtain the next frame’s viewpoint and render the corresponding tiles. Since neural rendering cannot precisely define the perceptual quality of individual tiles, the final rendered image results from the combined contributions of multiple overlapping tiles. After calculating the rendering loss across all stacked tiles, we define the perceptual quality of a single tile through occlusion weight. For rendering quality $Q_t^{\text {render}}$, we define it as:
\begin{equation}
    Q_t^{\text {render}}=\sum_{k=1}^{K_t} \sum_{r=1}^R \left[\mathrm{SSIM}_{fov} \cdot \Phi_{t,k} \cdot \Psi_{t,k} \right] \cdot x_{t, k, r},
\end{equation}
where $\operatorname{SSIM}_{fov}$ measures the structural similarity of the rendered image within the field of view. The occlusion weight $\Psi_{t, k}$ quantifies the contribution of tile $k$ in GoF $t$ based on its visibility and overlap with other tiles. The occlusion attenuation factor $\Psi_{t, k}$ is computed as:

%通过视角预测，我们可以提前获取下一时刻的用户视锥体，因此我们可以提前在指定视角下获取渲染图像，由于神经渲染无法准确定义单一切块的视觉质量，最终的渲染结果是由多个切块共同作用而成的，所以我们在计算多个叠加切块的渲染损失后采用遮挡权重来定义单个切块的渲染质量。其中遮挡衰减因子\Psi_{t,k}定义为：

\begin{equation}
    \Psi_{t,k}=\frac{1}{N_{t,k}} \sum_{i \in \mathcal{C}_{t,k}} \alpha_i \cdot e^{-\beta \cdot \frac{d_{t,k}}{d_{\max}}},
\end{equation}
with $N_{t, k}$ being the number of Gaussian primitives in tile $k$, $\mathcal{C}_{t, k}$ the set of all Gaussian primitives in tile $k$, and $d_{t, k}$ the Euclidean distance from center of tile $k$ to the viewpoint. The parameter $d_{\max }$ denotes the maximum distance of tile in the current viewport, and $\beta$ is attenuation coefficient.
%其中\alpha_i为切块k中第i个高斯基元的不透明度，{C}_{t,k}为切块k内高斯基元的集合，N_{t,k}为切块k内的高斯基元总数，d_{t,k}为切块中心到视点的欧氏距离，d_{\max }为视角最远点切块的欧氏距离，\beta为衰减系数。

%使用变形场进行编码的方式虽然可以很大限度地减少逐帧传输3DGS模型带来的数据量过大的问题，但即使是目前最新的动态高斯构建方法也需要33ms去解码目标帧的单个切块，考虑到在计算资源的限制下，大多数情况下需要分批次去解码所有切块，依然会因为解码时间造成卡顿从而极大影响用户的观看体验。但是提前构筑所有帧并进行传输也是不现实的，因为3DGS中每个高斯基元光是颜色表示就需要48个参数，相较于点云只需要6个参数来表示xyz以及rgb，3DGS视频的单帧大小甚至远超点云视频的单帧大小，如此巨大的数据量显然不适合逐帧传输。因此我们对每个切块设置了两种传输模式，一种是使用变形场进行编码并在传输至用户端后再进行解码的模式（encoded tiles），另一种模式为直传重构的目标切块(reconstructed tiles)。 
While encoding with deformation fields can significantly reduce the excessive data volume caused by frame-by-frame transmission of 3DGS models, even the current state-of-the-art dynamic Gaussian reconstruction methods require 33 ms to decode a single tile per target frame, severely impacting user viewing experiences in video streaming. However, pre-transmitting all frames is impractical because each Gaussian primitive in 3DGS requires as many as 48 parameters for color representation, whereas each point in point clouds inherently contains only 6 parameters (3 for position and 3 for RGB). Consequently, the per-frame size of 3DGS video far exceeds that of point cloud video, making frame-by-frame transmission infeasible. Therefore, we define two transmission modes for each tile: (1) Encoded tiles: Tiles encoded using deformation fields and decoded client-side after transmission. (2) Reconstructed tiles: Tiles bypass client-side processing entirely by pre-reconstructing geometric details through deformation fields and transmitting only finalized decoded data.
%对于GoF\t我们可以将解码时间 T_t^D定义为：
For each GoF $t$, we formalize its decoding time $T_t^D$ as:
\begin{equation}
    T_t^D=\frac{\sum_k^{K_t} e_{t, k} \times \sum_r^R\left(\varphi_{t, k, r} \times x_{t, k, r} \times f_{t, k}\right)}{C},
\end{equation}
where $e_{t, k} \in\{0,1\}$ indicates the transmission mode of tile $k$ (1: encoded mode, 0: reconstructed mode). $\varphi_{t, k, r}$ denotes decoding time per tile, linearly proportional to its Gaussian primitive count. $f_{t, k}=\mathbb{I}\left(v_{t, k}>0\right)$ acts as a viewport indicator function. $C$ represents the client's CPU core count enabling parallel decoding.
%其中e_{t, k}表示GoF\t中切块\k的传输形式，当其为1时表示以编码形式进行传输，当其为0时表示以重构形式传输切块。\varphi_{t, k, r}为切块的解码时间，其数值大体可按照切块内的高斯基元数进行线性求解。f_{t, k}是判断切块是否在FoV内的变量，当v_{t, k}为0时f_{t, k}也为0，当v_{t, k}大于0时f_{t, k}值为1。C为设备的CPU核心数量，表示能同时进行切块解码操作的线程数。

%对于GoF\t的传输时间 T_t^S我们将其定义为编码切块传输时间T_t^E与重构切块的传输时间T_t^R的和：
The total transmission time $T_t^S$ comprises encoded tile transmission $T_t^E$ and reconstructed tile transmission $T_t^R$:
\begin{equation}
    T_t^S=T_t^R+T_t^E,
\end{equation}
with each component calculated as:
\begin{equation}
    T_t^R=\frac{\sum_{k=1}^{K_t}\left[\left(1-e_{t, k}\right) \times \sum_{r=1}^R\left(S_{t, k, r}^E \times x_{t, k, r} \times f_{t, k}\right)\right]}{B_t},
\end{equation}
\begin{equation}
    T_t^E=\frac{\sum_{k=1}^{K_t}\left[e_{t, k} \times \sum_{r=1}^R\left(S_{t, k, r}^R \times x_{t, k, r} \times f_{t, k}\right)\right]}{B_t},
\end{equation}
where $S_{t, k, r}^E$ and $S_{t, k, r}^R$ denote the data sizes of encoded/reconstructed tiles respectively, and $B_t$ is the available network bandwidth.
%其中S_{t, k, r}^E和S_{t, k, r}^R分别代表GoF\t内所有以编码形式传输的切块k的数据量以及以重构形式传输的切块k的数据量。B_t表示流式传输GoF\t时的网络带宽。

%从客户端请求到所有数据被存储在buffer中的时间可以被定义为：
The end-to-end latency from client request to buffer readiness is:
\begin{equation}
    T_t^U=T_t^E+\max \left(T_t^R, T_t^D\right).
\end{equation}
This formulation reflects the pipelined transmission strategy: encoded tiles are transmitted first while reconstructed tiles and decoding operations proceed concurrently.
%这是由于我们需要先发送以编码形式传输的切块，并在解码的同时传输重构形式的切块。

%最终我们可以按照transmission time和playback buffer dynamics来计算stall time与stall frequency：
The stall time $P_t^d$ and stall frequency $P_t^f$ are derived from playback buffer dynamics:
\begin{equation}
    P_t^d=\left(T_t^U-T^I-\mathbb{L}_{t-1}\right)_{+},
\end{equation}
\begin{equation}
    P_t^f=\left\{\begin{array}{l}
1, T_t^U-T^I-\mathbb{L}_{t-1}>0 \\
0, \text { else }
\end{array}\right.,
\end{equation}
where $T^I$ is the GoF playback duration and $\mathbb{L}_{t-1}$ represents the previous buffer occupancy.
%其中T^I为单个GoF的playback time，\mathbb{L}_{t-1}为前一时刻的缓冲区长度。

The dynamic nature of 3DGS video streaming, characterized by viewport-dependent rendering, heterogeneous network conditions, and computational constraints, necessitates a principled framework for joint bitrate adaptation and transmission mode optimization. We formulate this challenge as a meta-reinforcement learning (meta-RL) task, where an agent iteratively interacts with the streaming environment to learn a policy that maximizes the expected long-term QoE. At each decision epoch, the agent observes a state encoding viewport dynamics, network bandwidth variability, and client-side resource utilization. It then selects actions that jointly determine two critical parameters: the encoding mode (encoded or reconstructed) and the quality level for each spatial tile.

%状态：实时网络带宽B_t，当前缓冲区的占用率，预测的FoV视点以及覆盖切块的可见比例v_{t, k}，当前切块空间影响因子\Phi_{\text {spatial }}(k)以及遮挡衰减因子\Psi_{\text {occlusion }}(k)。
The state vector integrates spatiotemporal viewport dynamics, network conditions, and content characteristics to guide adaptive decision-making. It is formally defined as:
\begin{equation}
    s_t=\left(\vec{v}_t, \vec{s}_t, \vec{B}_t, \mathbb{L}_{t-1}, \mathcal{C}_t, \sum e_{t, k} N_{t, k}\right),
\end{equation}
where $\vec{v}_t$ captures the 3D trajectory of the viewport center over five consecutive frames, directly linked to the tile visibility ratio $v_{t, k}$. The saliency vector $\vec{s}_t$ provides per-tile visual prominence weights $s_{t, k}$, which modulate the spatial influence factor $\Phi_{t, k}$ in the QoE model. Network status $\overrightarrow{B_t}$ characterizes bandwidth availability through its instantaneous value and temporal variance, governing the denominator in transmission time equations $T_t^E$ and $T_t^R$. Buffer occupancy $\mathbb{L}_{t-1}$ serves as a critical constraint for stall time calculation $P_t^d$, while the content descriptor $\mathcal{C}_t$ encodes scene-specific dynamics affecting Gaussian primitive density $N_{t, k}$. The decoding load term $\sum e_{t, k} N_{t, k}$ quantifies computational overhead by aggregating Gaussian primitives in encoded tiles, directly influencing $T_t^D$.

%动作：切块传输版本选择e_{t, k} \in\{0,1\}，0为编码版本，1为解构版本。切块质量等级选择x_{t, k, r} \in\{1, \ldots, R\}。动作的目标是最大化长期累积奖励QoE。
The action space comprises two interdependent decisions: transmission mode selection and quality level allocation. Each tile’s transmission mode $e_{t, k} \in\{0,1\}$ determines whether it is encoded (1) or reconstructed (0), subject to the global bandwidth constraint:
\begin{equation}
    \sum\left[e_{t, k} S_{t, k, r}^E+\left(1-e_{t, k}\right) S_{t, k, r}^R\right] x_{t, k, r} \leq B_t T^I.
\end{equation}
Concurrently, the quality selection variable $x_{t, k, r} \in\{0,1\}$ adheres to the exclusivity constraint $\sum_{r=1}^R x_{t, k, r}=1$, ensuring exactly one quality level is chosen per tile. This dual-action structure enables joint optimization of bandwidth utilization and rendering quality.

%奖励：即时奖励为QoE，由于视频在GoF之间使用不同关键帧加初始模型进行构建，所以闪烁或突变的情况会更严重，所以需要引入平滑性约束S_t，最终的累计奖励r_t为：
To address quality fluctuations between Groups of Frames (GoFs) caused by keyframe-based 3DGS model transitions, we enhance the QoE-driven reward function with a temporal smoothness penalty. The revised reward formulation integrates geometric consistency and rendering continuity across consecutive GoFs:
\begin{equation}
    r_t=\lambda\left(\alpha Q_{\mathrm{geo}}+(1-\alpha) Q_{\mathrm{render}}\right)-\mu P_t^d-\sigma P_t^f-\eta S_t,
\end{equation}
\begin{equation}
    S_t=\delta \cdot\left\|Q_{\text {geo }}^t-Q_{\text {geo }}^{t-1}\right\|_2+(1-\delta) \cdot\left\|Q_{\text {render }}^t-Q_{\text {render }}^{t-1}\right\|_2.
\end{equation}
The smoothness term $S_t$ penalizes abrupt quality variations through two components, $Q_{\text {geo }}^t$ and $Q_{\text {render }}^t$ denote the geometric and rendering quality metrics of GoF $t$, respectively. The balance factor $\delta \in[0,1]$ adapts to scene dynamics, prioritizing geometric stability for fast-motion sequences and rendering consistency for viewport rotations. The unified formulation preserves the core QoE optimization objectives while explicitly mitigating inter-GoF artifacts inherent to 3DGS streaming architectures.

To address the scarcity of high-quality 3DGS training data and enable robust generalization across diverse streaming scenarios, we integrate model-agnostic meta-learning (MAML) into the reinforcement learning framework. During meta-training, the agent is exposed to a distribution of tasks $\mathcal{T}_i$, each defined by a unique combination of bandwidth profiles and content dynamics. For each task, the policy learns to optimize the composite reward $r_t$ while adapting to two critical dimensions: (1) Bandwidth variability: Simulated throughput fluctuations mimic real-world network conditions, requiring dynamic trade-offs between encoded and reconstructed tiles. (2) Content Heterogeneity: Scene-specific Gaussian distributions demand adaptive spatial weighting of $\Phi_{t, k}$ and $\Psi_{t, k}$. 

The meta-learning objective trains an initial policy $\pi_\theta$ that can rapidly adapt to unseen tasks with minimal fine-tuning episodes. This is achieved through bi-level optimization: 

Inner loop: For task $\mathcal{T}_i$, perform gradient updates on a support set of streaming trajectories to minimize:
\begin{equation}
    \mathcal{L}_{\text {inner }}^i=-\mathbb{E}_{(s, a, r) \sim \mathcal{D}_{\text {supp }}}\left[\sum_{\tau=0}^H \gamma^\tau r_\tau\right]+\xi \cdot \mathrm{KL}\left(\pi_{\theta^{\prime}}| | \pi_\theta\right),
\end{equation}
where $\gamma$ is the discount factor and $\xi$ regularizes policy divergence.

Outer loop: Update the meta-policy $\theta$ by evaluating adapted policies $\pi_{\theta^{\prime}}$ on query sets from all tasks:
\begin{equation}
    \theta^*=\arg \min _\theta \sum_{\mathcal{T}_i} \mathcal{L}_{\text {outer }}^i\left(\pi_{\theta^{\prime}}\right),
\end{equation}
\begin{equation}
    \theta^{\prime}=\theta-\alpha \nabla_\theta \mathcal{L}_{\text {inner }}^i.
\end{equation}
To encode task-specific characteristics, we introduce a learnable embedding $z_{\mathcal{T}}=\operatorname{Enc}\left(\mathcal{C}_t \oplus \vec{B}_t\right)$, where $\mathcal{C}_t$ captures scene dynamics via Gaussian centroid displacements obtained from each tile’s deformation field, and $\vec{B}_t$ encodes bandwidth statistics. This embedding modulates the policy network through feature-wise linear modulation (FiLM) layers:
\begin{equation}
    \operatorname{FiLM}(h)=\beta\left(z_{\mathcal{T}}\right) \odot h+\gamma\left(z_{\mathcal{T}}\right),
\end{equation}
where $h$ denotes hidden layer activations, and $\beta$, $\gamma$ are generated by multi-layer perceptrons. Crucially, the QoE weighting parameters $\lambda, \mu, \sigma, \eta$ are dynamically generated as $\operatorname{MLP}\left(z_{\mathcal{T}}\right)$ rather than fixed, allowing automatic prioritization of geometric fidelity, rendering quality, stall avoidance, and temporal smoothness based on current task requirements. For instance, in bandwidth-constrained scenarios, the meta-policy learns to increase $\mu$ (stall penalty weight) while decreasing $\lambda$ (quality emphasis), whereas viewport-unstable tasks upweight $\eta$ (smoothness penalty).

\begin{figure}[h!]
    \centering
    \begin{subfigure}[t]{0.19\textwidth} 
        \centering
        \includegraphics[width=\textwidth]{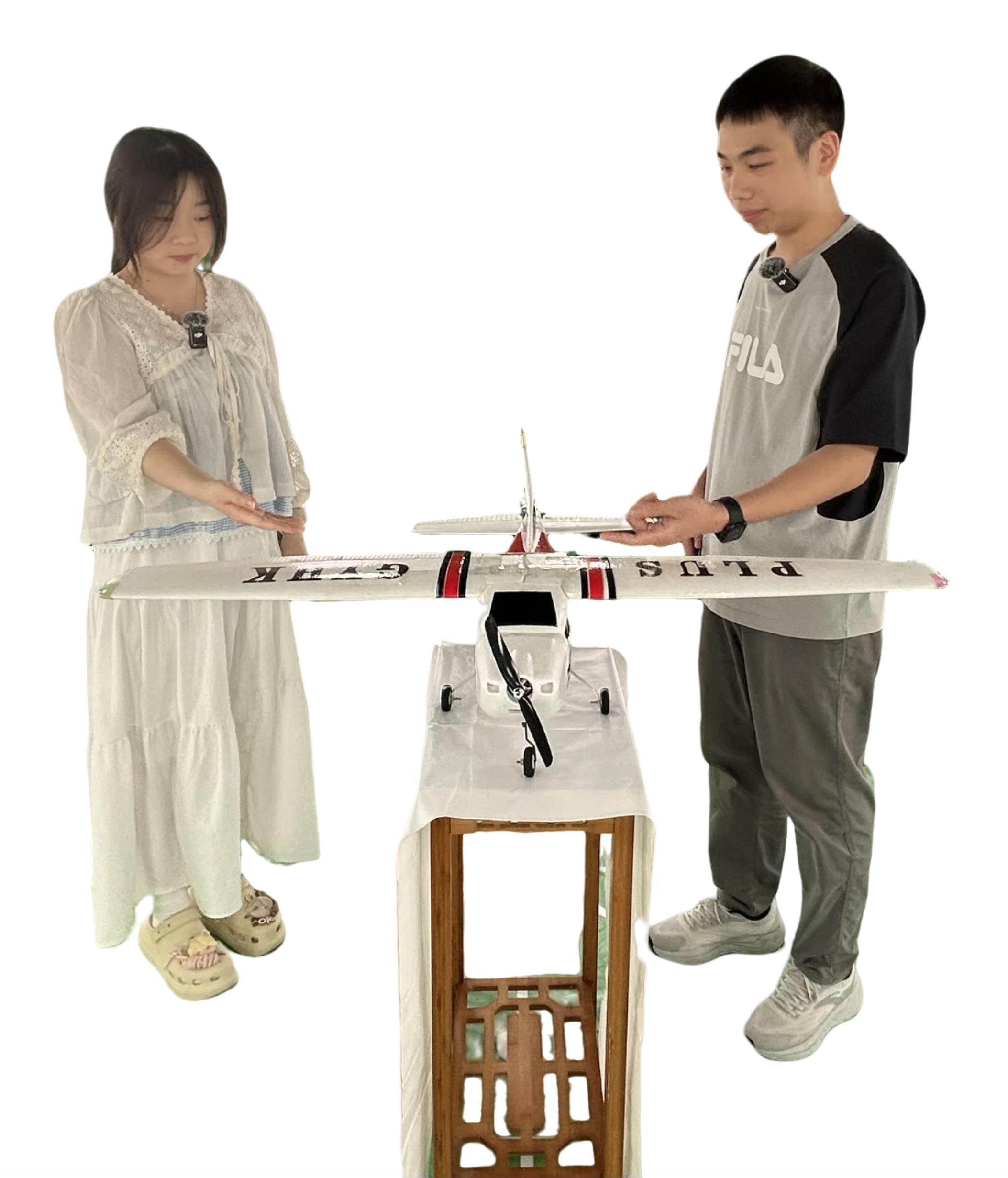}
        \caption{}
        \label{dataset1}
    \end{subfigure}
    \hfill 
    \begin{subfigure}[t]{0.19\textwidth}
        \centering
        \includegraphics[width=\textwidth]{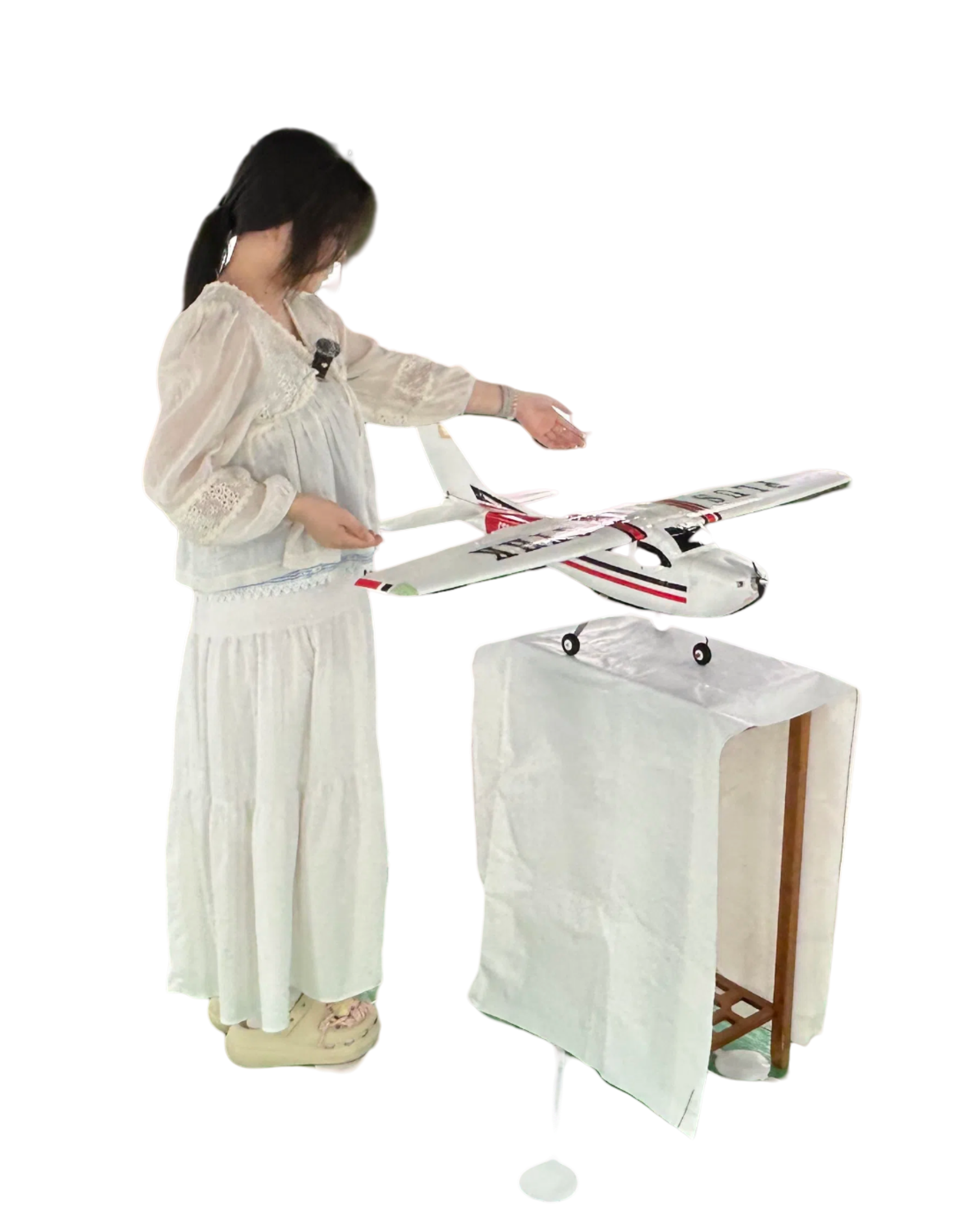}
        \caption{}
        \label{dataset2}
    \end{subfigure}

    \vspace{0cm}
    
    \begin{subfigure}[t]{0.19\textwidth}
        \centering
        \includegraphics[width=\textwidth]{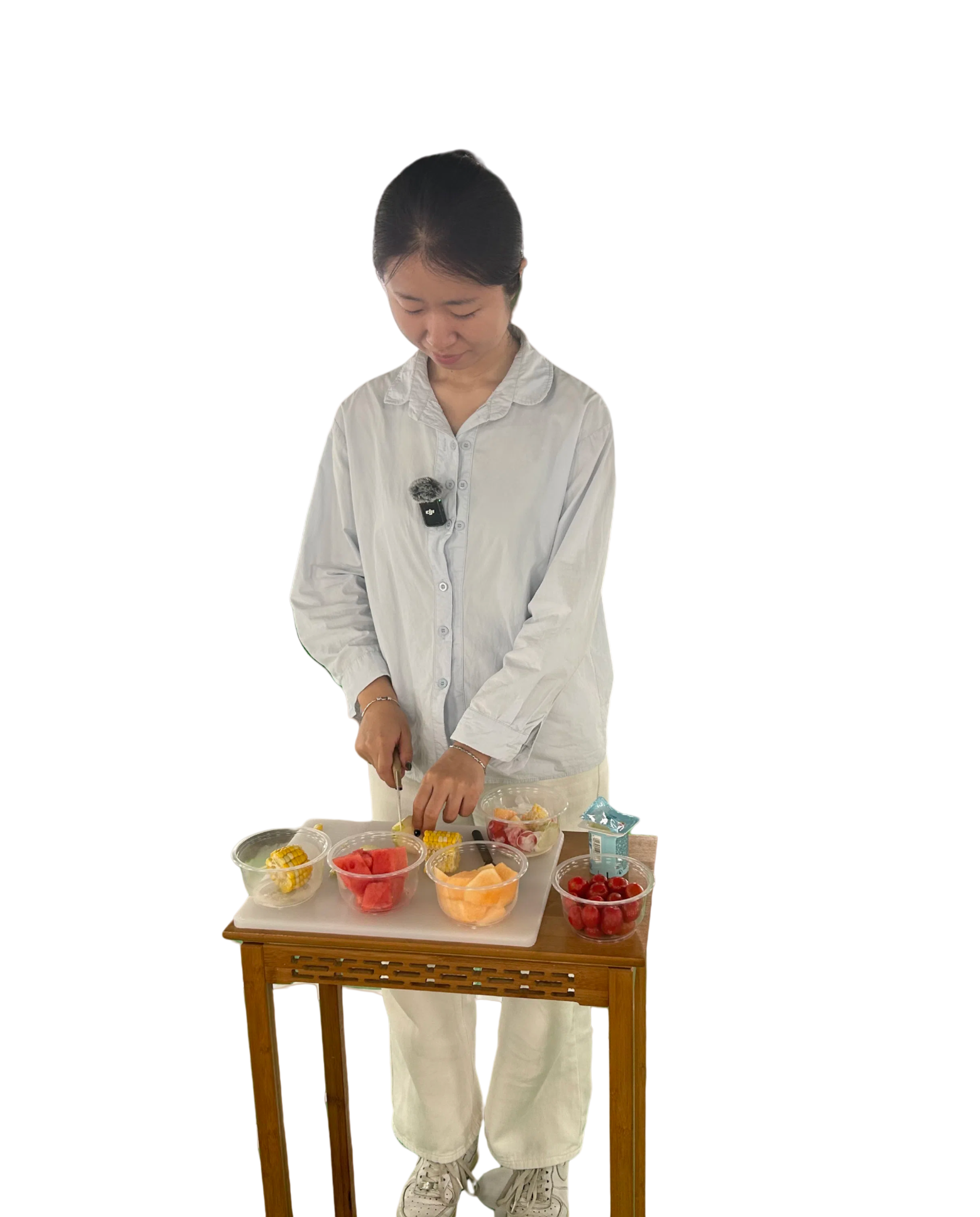}
        \caption{}
        \label{dataset3}
    \end{subfigure}
    \hfill
    \begin{subfigure}[t]{0.19\textwidth}
        \centering
        \includegraphics[width=\textwidth]{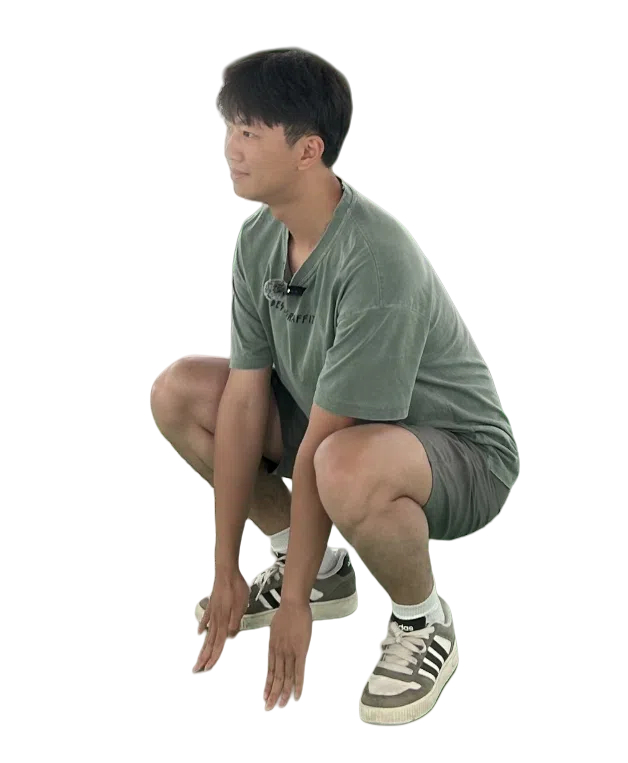}
        \caption{}
        \label{dataset4}
    \end{subfigure}

    \caption{Partial sequence demonstrations from our dataset.}
    \label{Dataset}
\end{figure}

\section{EXPERIMENT}
%$rgbd.yam1
\subsection{Experimental Setup}
%Video Source：我们使用公开数据集HiFi4G、DNArandering中的部分序列作为3DGS视频来源。同时为了扩展元学习算法在不同场景下的泛用性以及提高用户在观看视频时行为的可控性（有更统一的head trajectory），我们录制了一些新的视频序列，这些序列中包含了更复杂的场景（例如多人交互、遮挡场景、更丰富的剧情）以及更长的视频时长。所有的序列通过本研究的重建与切块方案进行预先处理。我们将这些序列按照视频类型进行分类，以此来验证我们的方案在面对不同种类视频时的效果。策略主要有两个，一个是按照视频观看时是否有视角限制，即是否可以进行无死角观看区分为受限序列和不受限序列，因为这会严重影响视角预测时用户的视角信息，并直接影响传输时的决策。第二个则是按照视频中人物的运动幅度分为高动态序列和低动态序列，动态程度直接与切块的显著性挂钩，且会影响不同GoF间的质量变动情况。一般来说高动态的序列在GoF之间会存在更严重的质量突变从而导致画面闪烁。帧率设置为30Hz，每个视频被编码为5个质量版本和两个传输版本。
\textbf{Video Source}: The 3DGS video sources in our experiments comprise selected sequences from the publicly available HiFi4G \cite{jiang2024hifi4g} and DNA-Randering \cite{cheng2023dna} datasets. To enhance the cross-scenario generalizability of the meta-learning algorithm and improve controllability over user viewing behavior by establishing more consistent head trajectories, we captured additional video sequences featuring complex scenarios including multi-person interactions (Figures \ref{dataset1} and \ref{dataset2}), occlusion situations (Figure \ref{dataset3}), and narrative-rich environments with extended durations (Figure \ref{dataset4}). All sequences undergo preprocessing through our proposed reconstruction and tiling framework. Videos were divided into constrained (CS)/unconstrained (UC) sequences based on viewing perspective restrictions, as this classification directly impacts viewpoint prediction accuracy and transmission decision-making. Additionally, sequences were labeled as high-dynamic (HD) or low-dynamic (LD) according to subject motion amplitude , as dynamic sequences exhibit more pronounced quality fluctuations across Group-of-Frames (GoFs) and increased risk of rendering artifacts. All videos were standardized to 30 Hz frame rate and encoded into five quality versions via multi-quality encoding, alongside two transmission versions (encoded vs. reconstructed tiles).

%FoV Trace：我们将所有处理完成的3DGS视频序列通过Unity环境导入meta quest3中，之后召集了50名志愿者通过VR设备观看这些序列并记录所有观看者的head trajectory。在此基础上我们修改了我们之前的视角预测工作\cite{livpformer}，使其适用于3DGS视频的视角预测。最终获取了每个序列的预测FoV trace，包括视点的空间位置(x,y,z)和用户头部的位姿(pitch, yaw, roll)。这些视角信息被用于辅助我们选择所要传输的切块以及对应的质量等级。
\textbf{FoV Trace}: The processed 3DGS video sequences were imported into Meta Quest 3 headsets via the Unity environment, with 50 participants recruited to view these sequences through VR devices while their head movement trajectories were systematically recorded. Building upon this empirical dataset, we refined our prior viewpoint prediction framework \cite{livpformer} to align with the unique characteristics of 3DGS video content. This adaptation yielded predicted FoV traces for each sequence, encompassing spatial viewpoint coordinates (x, y, z) and detailed head orientation parameters (pitch, yaw, roll). These comprehensive viewpoint metrics serve as critical inputs for optimizing tile selection and quality-level allocation during the streaming process, ensuring prioritized transmission of regions within predicted visual attention zones.

%Bandwidth Trace：我们采用真实环境下差异较大的几个网络条件来实现我们元学习的方法在面对不同网络环境时的适应能力。我们从4G trace数据集\cite{raca2018beyond}\cite{mei2020realtime}、5G trace数据集\cite{raca2020beyond}中采样了多条序列，并按照其带宽峰值与均值将其区分为4个不同的组，用于代表四种不同的通信场景。四组的范围如下：标准4G带宽环境（Std4G）：35 Mbps到90 Mbps，极端波动4G环境（Ext4G）：0 Mbps到150 Mbps，标准5G带宽环境（Std5G）：150 Mbps到600 Mbps，极端5G环境（Ext5G）：0 Mbps到1200 Mbps。
\textbf{Bandwidth Trace}: To rigorously evaluate the meta-learning algorithm's adaptability across heterogeneous network environments, we employed diverse real-world bandwidth patterns sampled from established 4G network trace datasets \cite{raca2018beyond}\cite{mei2020realtime} and 5G network trace dataset \cite{raca2020beyond}. These traces were systematically categorized into four distinct communication scenarios based on their peak bandwidth values and temporal stability characteristics: Standard 4G environment (Std4G) with sustained bandwidth ranging from 35 Mbps to 90 Mbps, Extreme 4G environment (Ext4G) exhibiting volatile fluctuations between 0 Mbps and 150 Mbps, Standard 5G environment (Std5G) maintaining stable connections from 150 Mbps to 600 Mbps, and Extreme 5G environment (Ext5G) demonstrating highly erratic patterns spanning 0 Mbps to 1200 Mbps.

\begin{figure}[h!]
    \centering
    
    \begin{subfigure}[t]{0.5\textwidth}  
        \centering
        \includegraphics[width=1\textwidth]{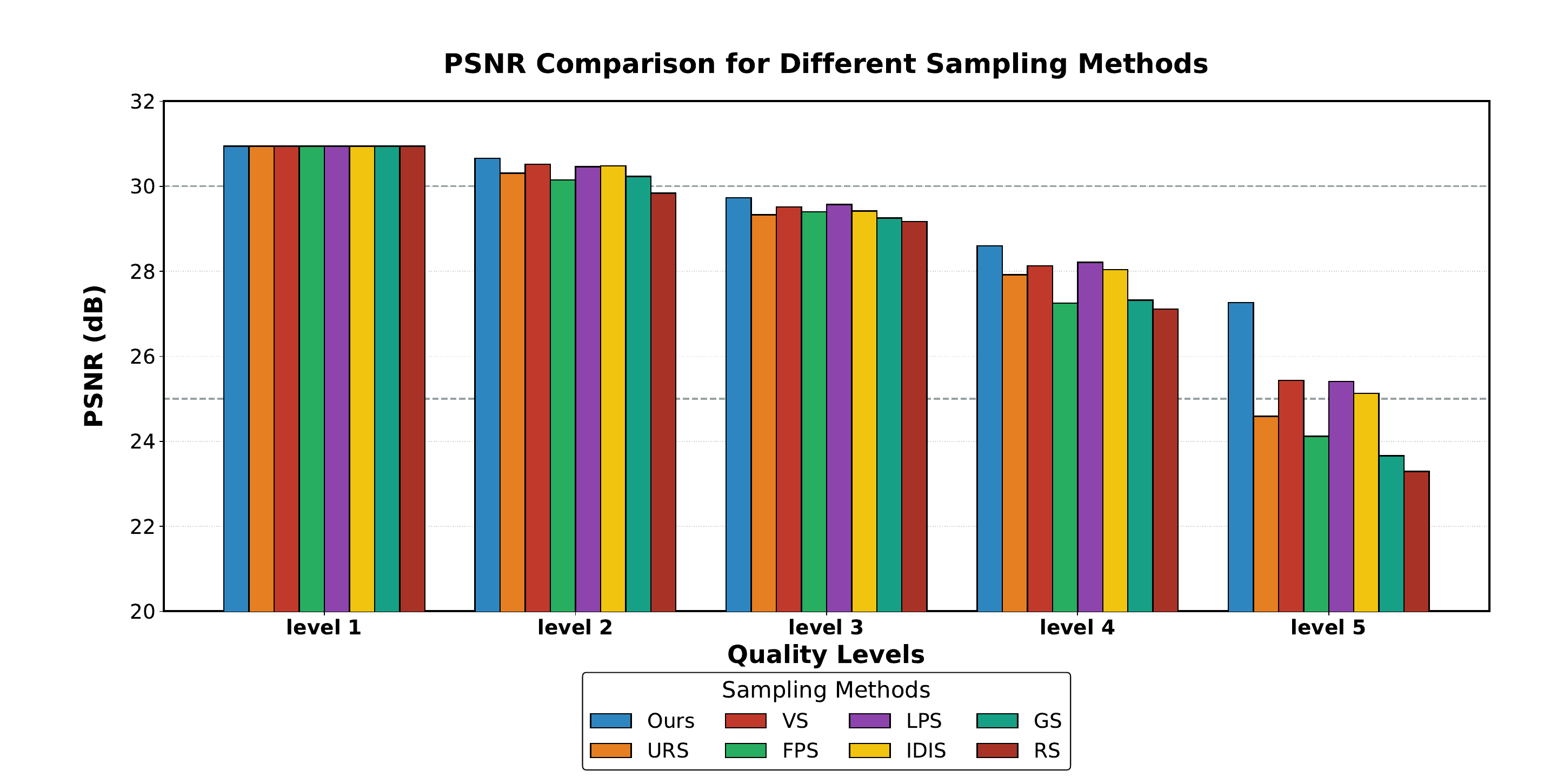}  
        \caption{}
        \label{sample1}
    \end{subfigure}
        %\vspace{0em}  
    \begin{subfigure}[t]{0.5\textwidth} 
        \centering
        \includegraphics[width=1\textwidth]{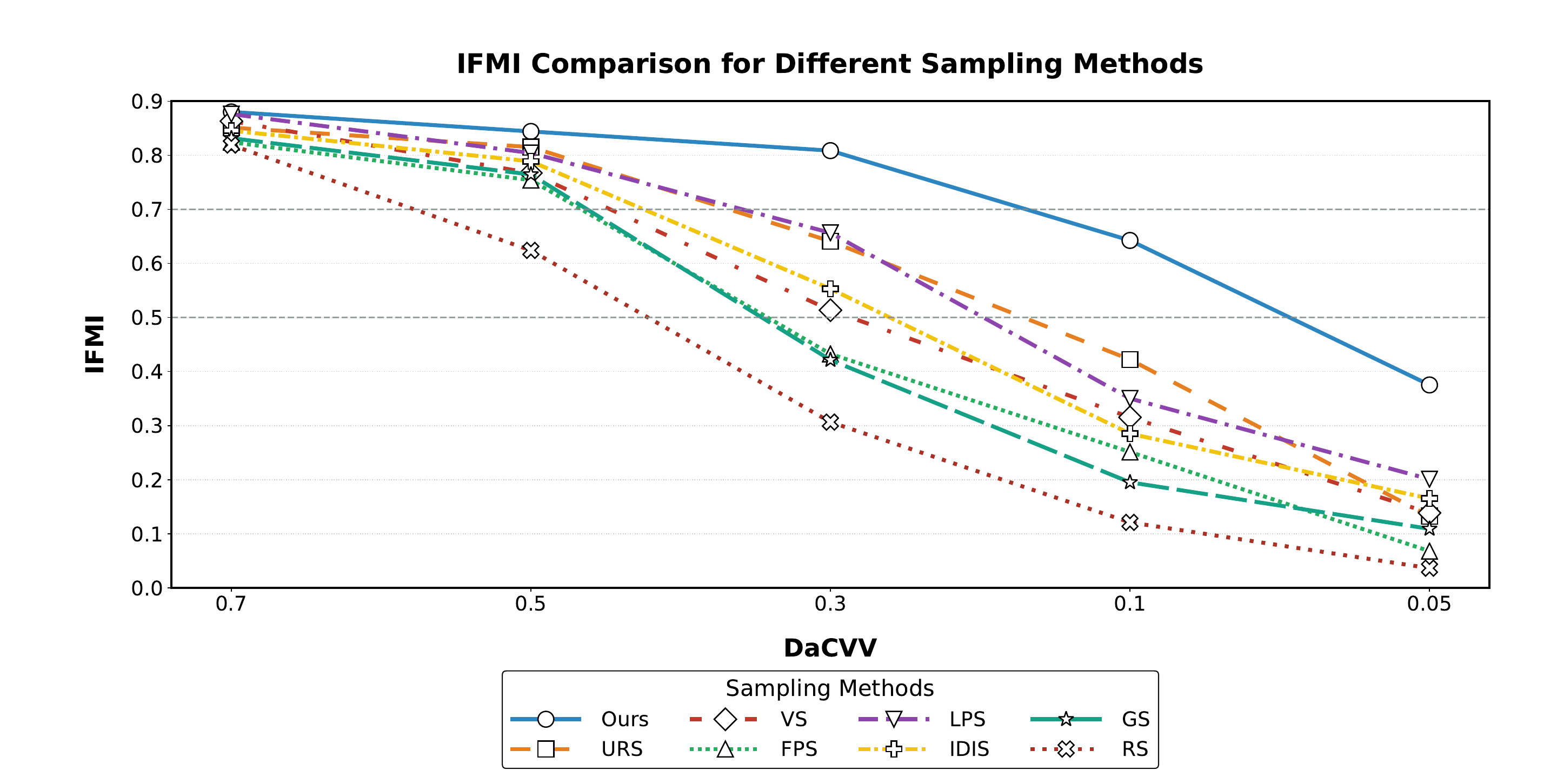}
        \caption{}
        \label{sample2}
    \end{subfigure}
    
    \caption{Comparison of different sampling and multi-quality compression methods.}
    \label{Result1}
\end{figure}

\begin{figure*}[h!]
    \centering

    \begin{subfigure}[t]{0.5\textwidth}
        \centering
        \includegraphics[width=\textwidth]{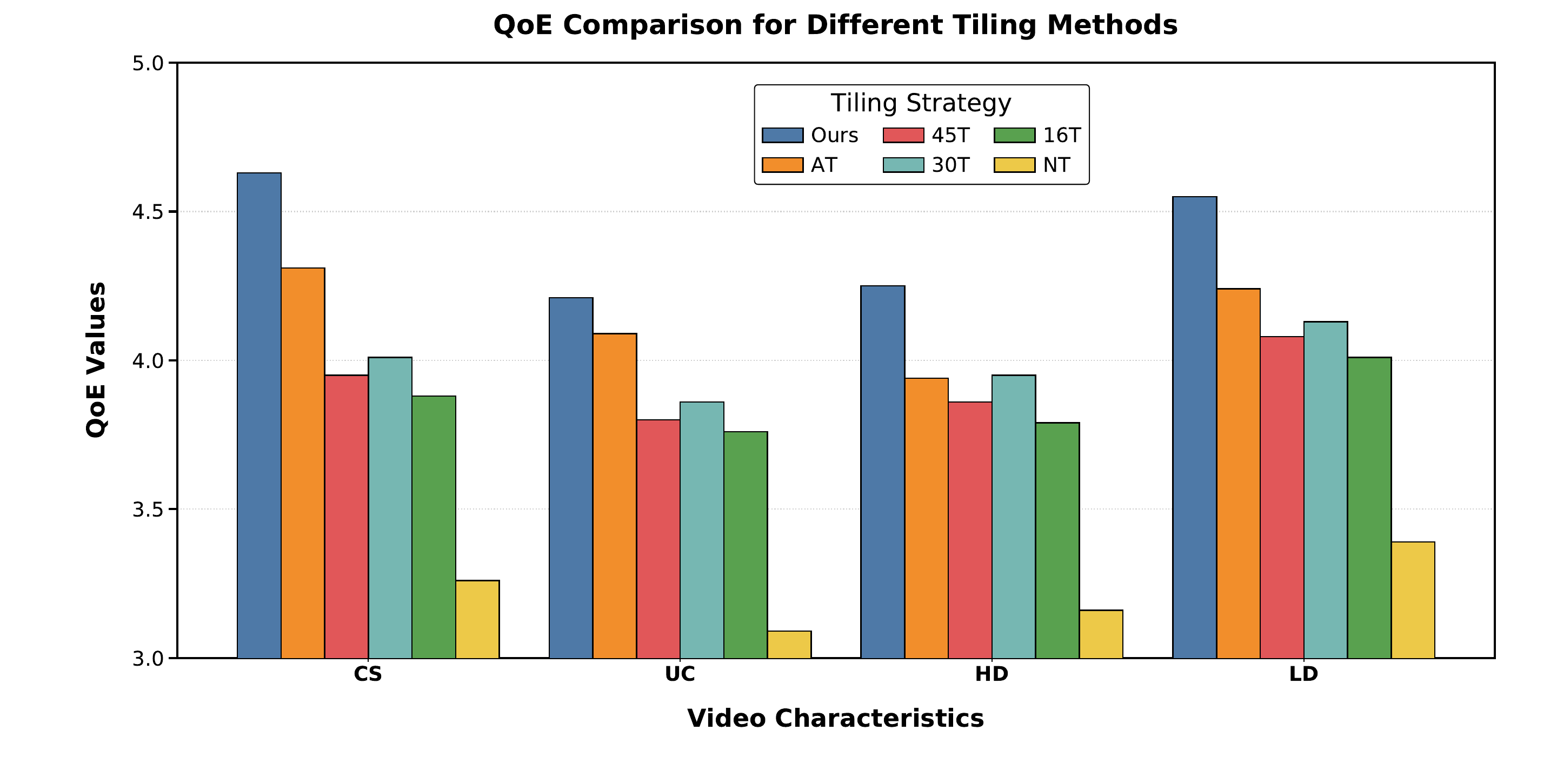}
        \caption{}
        \label{tile1}
    \end{subfigure}%
    \hfill
    \begin{subfigure}[t]{0.5\textwidth}
        \centering
        \includegraphics[width=\textwidth]{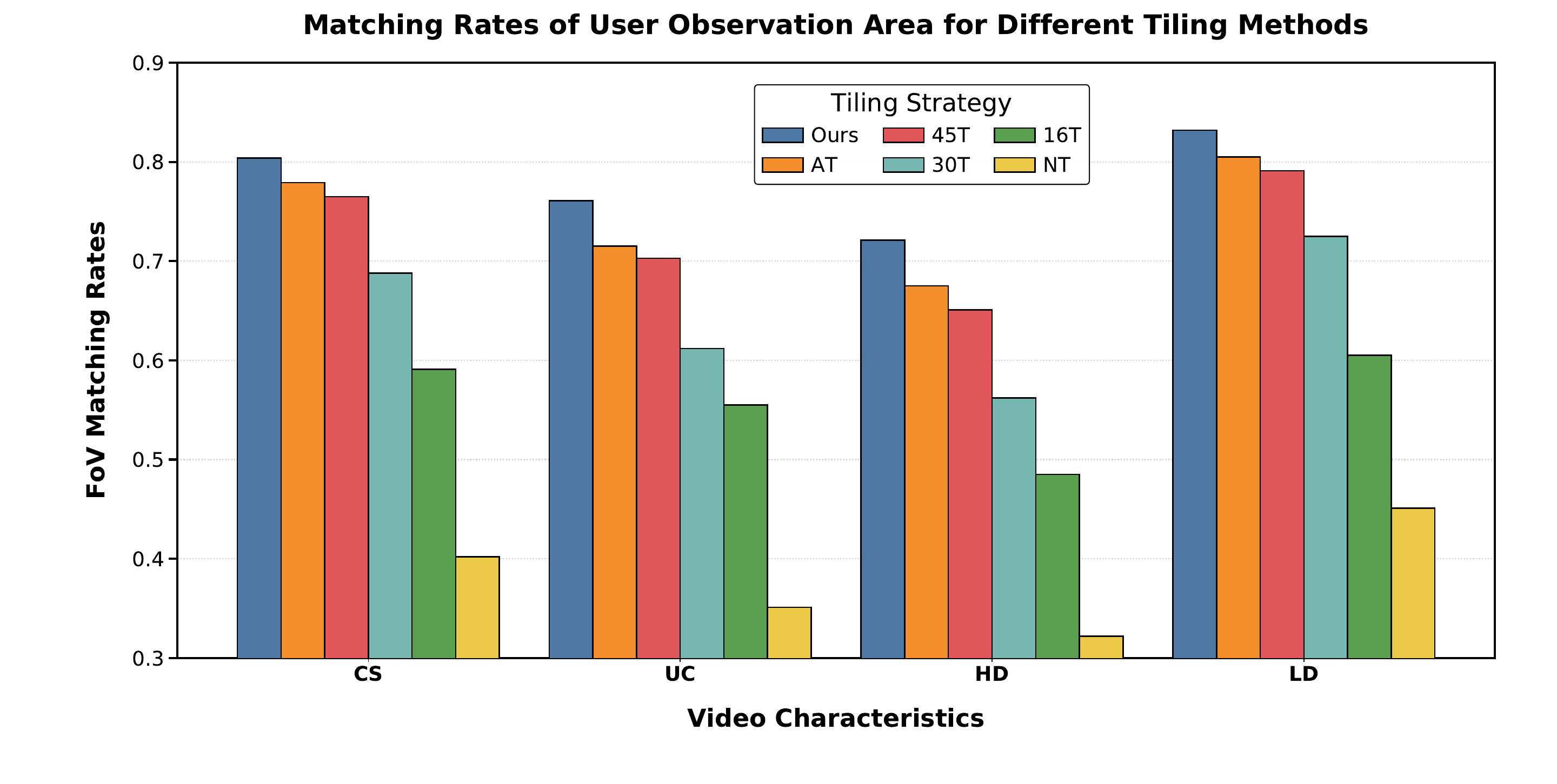}
        \caption{}
        \label{tile2}
    \end{subfigure}
\caption{Comparison of different tiling methods.}
    \label{Result2}
\end{figure*}

%Comparison Methods: 我们按照所提出方法的各个组成部分分别对比了一些体积视频中常见的解决方案，来验证我们方法在面对3DGS视频streaming时的有效性。1、特征提取前的采样以及切块的质量等级区分环节，我们将提出的基于渲染权重的采样方法与其他基于点和基元的sampling方法进行了对比，以此来验证我们方法在保留更重要高斯基元时的有效性。2、我们对比了提出的自适应切块方案与其他切块方案在性能方面的差距，包含对于QoE的贡献以及传输切块和实际用户FoV之间匹配率。3、我们对比了所提出的基于元学习的ABR算法与其他体积视频ABR算法在面对不同带宽环境时的泛用性，以及在面对小样本新环境时的学习能力。
%\textbf{Comparison Methods}: Our experimental evaluation systematically compares each component of the proposed framework against established volumetric video streaming solutions to validate its effectiveness for 3DGS content delivery. For the pre-processing stage involving feature sampling and quality-tier partitioning, we benchmark our rendering-weight-based sampling strategy against conventional point-based and primitive-oriented sampling methods, demonstrating superior preservation of perceptually critical Gaussian primitives. The adaptive tiling mechanism is rigorously evaluated against alternative tiling schemes through quantitative analysis of Quality-of-Experience (QoE) improvements and spatial matching rates between transmitted tiles and actual user Field-of-View regions. Furthermore, we assess the proposed meta-learning ABR algorithm's cross-environment generalization capabilities by comparing its performance with state-of-the-art volumetric video ABR solutions under diverse bandwidth conditions, particularly emphasizing its few-shot adaptation efficiency when encountering novel network environments.

\begin{figure*}[h!]
    \centering

    \begin{subfigure}[t]{0.5\textwidth}
        \centering
        \includegraphics[width=\textwidth]{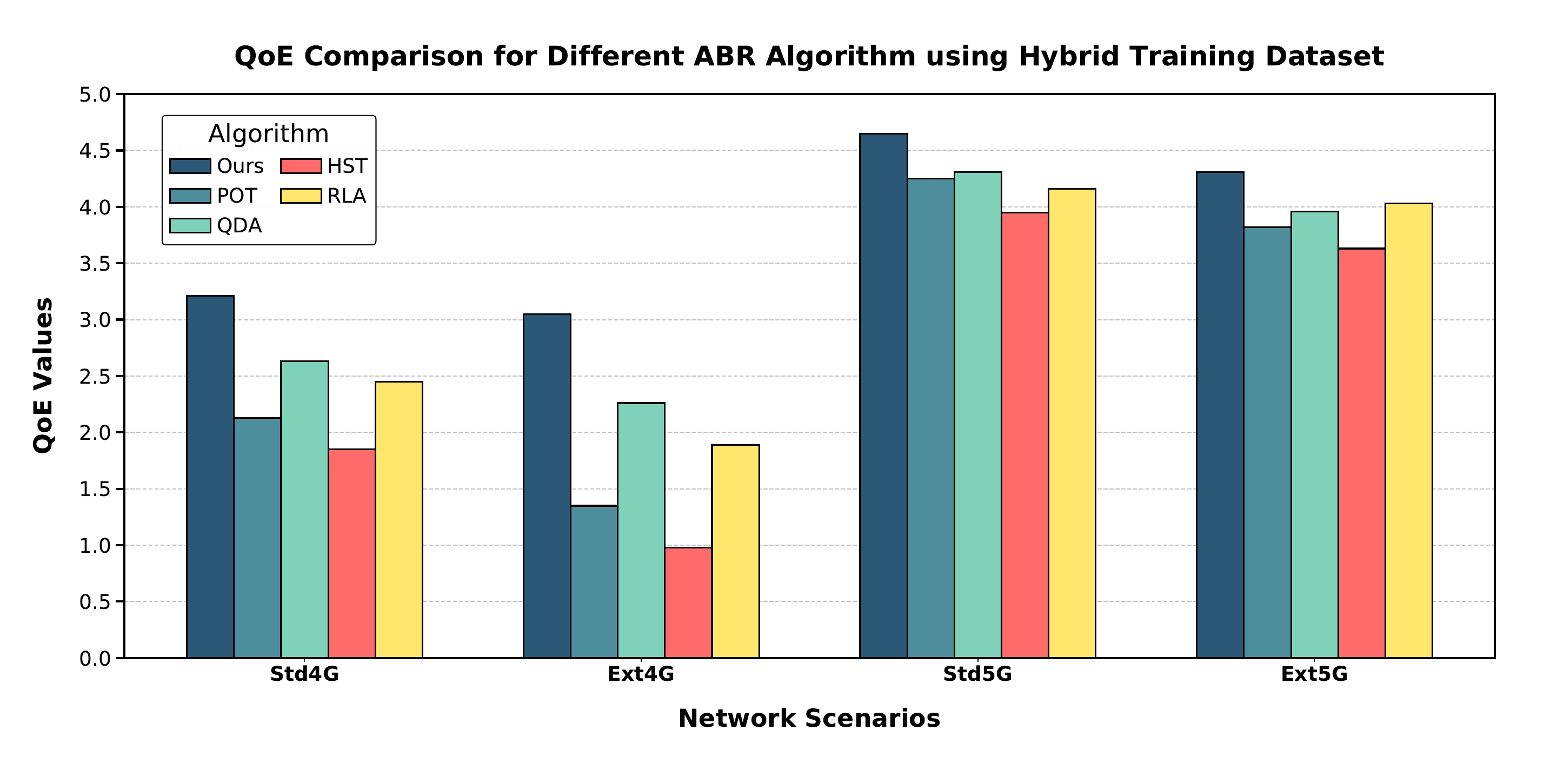}
        \caption{}
        \label{ABR1}
    \end{subfigure}%
    \hfill
    \begin{subfigure}[t]{0.5\textwidth}
        \centering
        \includegraphics[width=\textwidth]{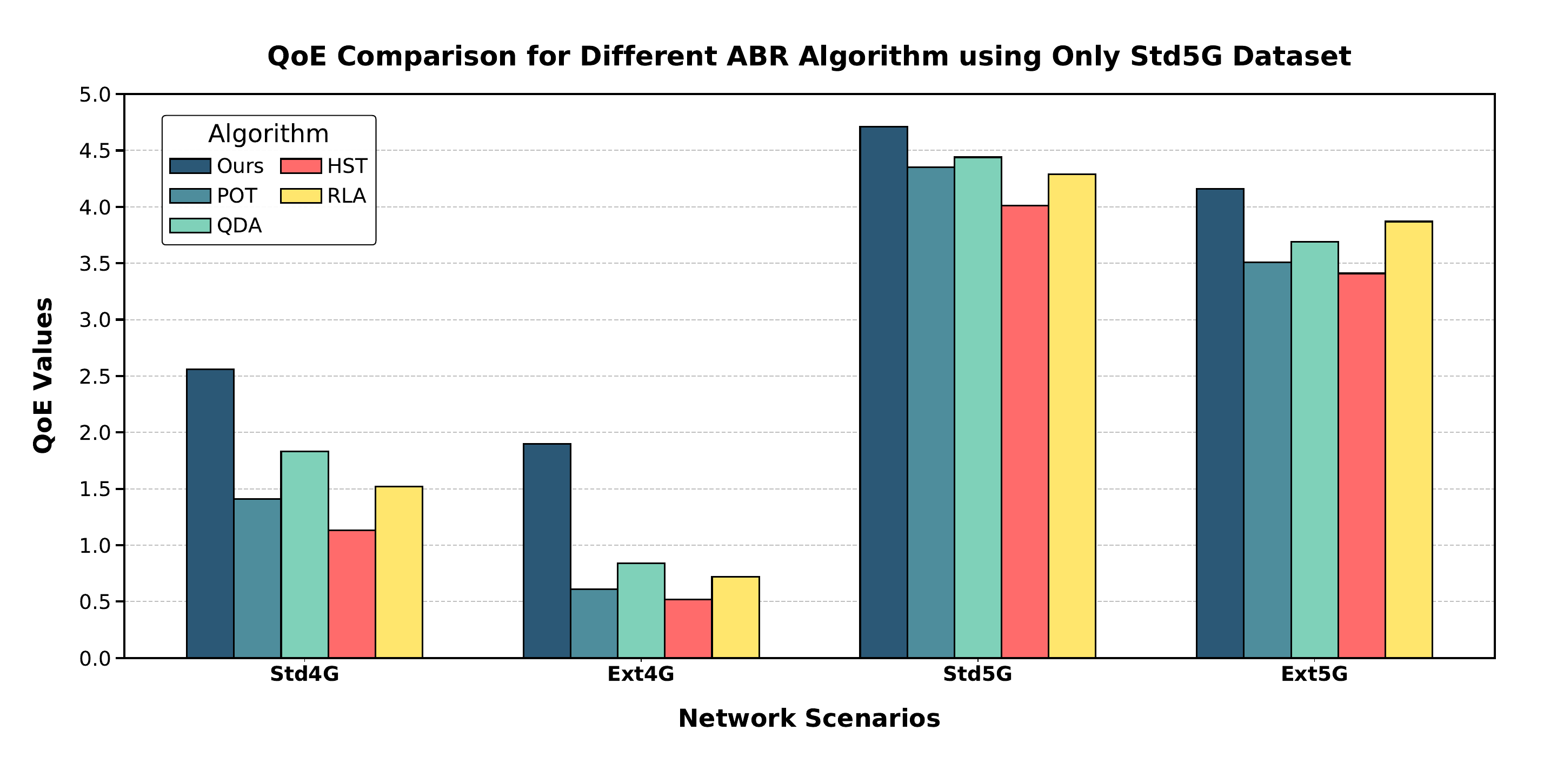}
        \caption{}
        \label{ABR2}
    \end{subfigure}
    
\vspace{0.2cm} 

    \begin{subfigure}[t]{0.5\textwidth}
        \centering
        \includegraphics[width=\textwidth]{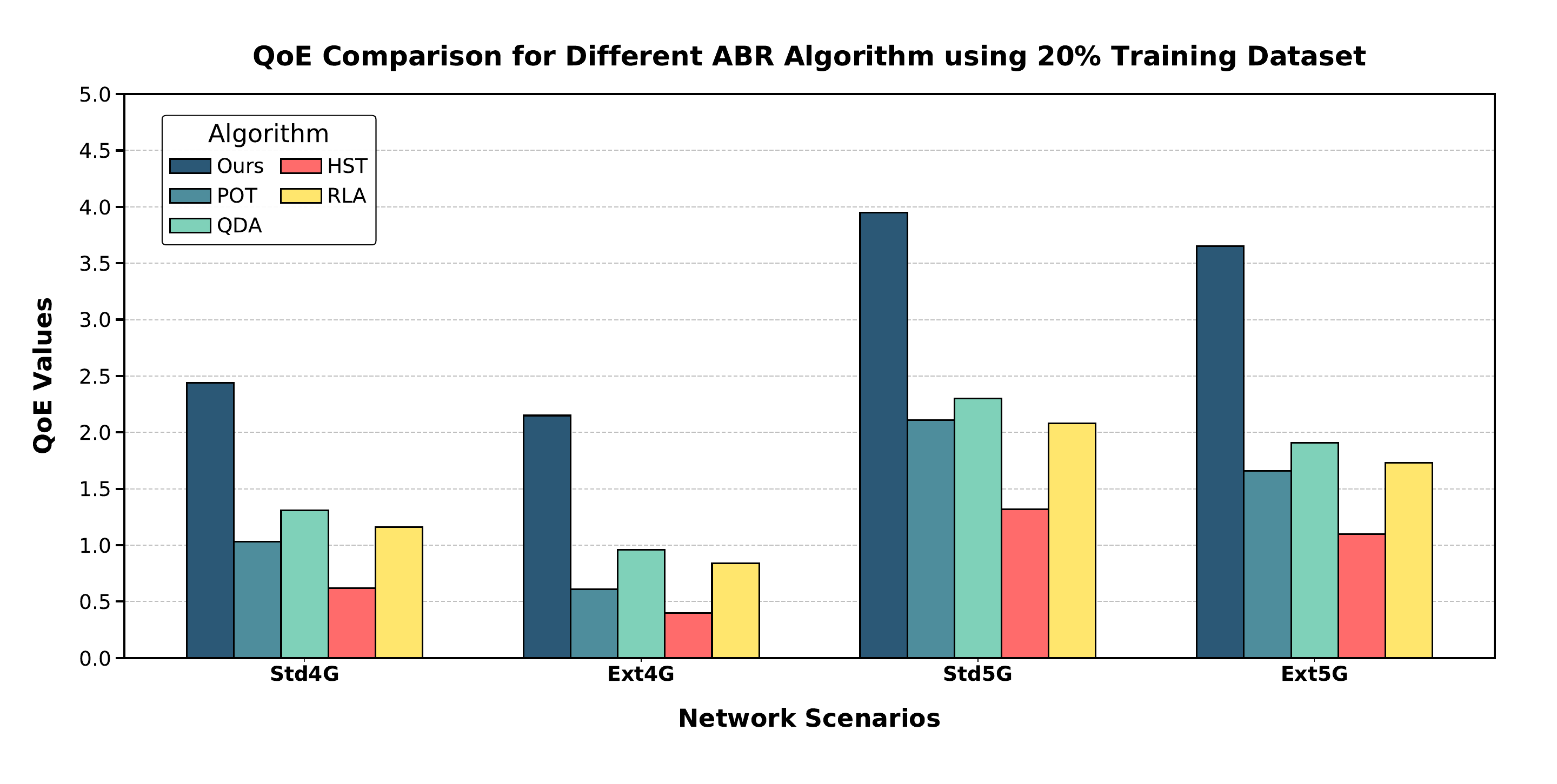}
        \caption{}
        \label{ABR3}
    \end{subfigure}%
    \hfill
    \begin{subfigure}[t]{0.5\textwidth}
        \centering
        \includegraphics[width=\textwidth]{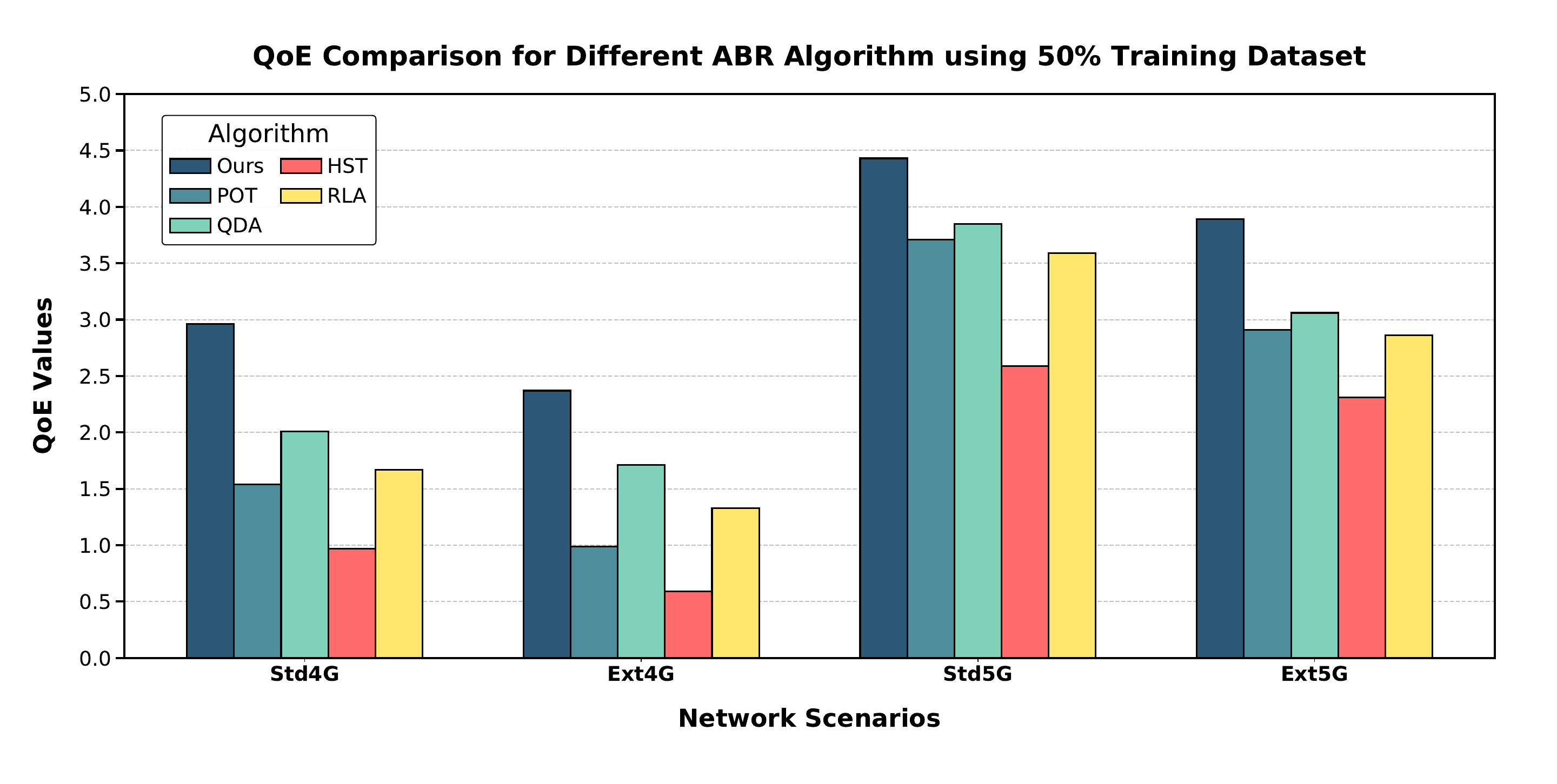}
        \caption{}
        \label{ABR4}
    \end{subfigure}

\caption{Comparison of different ABR algorithms.}
    \label{Result3}
\end{figure*}

\subsection{Experimental Results}
%\textbf{Sampling and multi-quality compression method}: 为了验证我们基于高斯渲染权重的采样和压缩方案能够保留那些对视频更重要的高斯基元，我们与其他几种采样基线方法在视频质量与帧间映射强度（inter-frame mapping intensity (IFMI)）上进行了对比。视频质量我们采用渲染后的2D图像与训练图片在同视角下的PSNR作为指标。IFMI我们通过设置不同的distance and color variation values（DaCVV）阈值来对比采样后各个GoF之间的时序关联程度，基线方法包括URS VS FPS LPS IDIS GS RS。
%\ref{sample1}中展示了我们将切块压缩到不同质量等级时的PSNR损失情况，其中每个等级的压缩百分比是固定的，可以明显地看出我们的方法在压缩的同时较好地保存了模型的渲染质量。而且压缩比越大我们的方法保留质量的效果越明显，这是由于当高斯基元缩减到一定数量时会导致渲染画面质量发生突变，此时剩余的高斯基元不足以支撑完整的渲染画面，因为我们的方法保留了对于渲染而言更重要的高斯基元，因此可以缓解画面的突变。不同采样方法在给定DaCVV阈值时的IFMI变化情况如\ref{sample2}所示，同样的我们的方法实现了更好的帧间关系映射。这是由于3DGS与点云不同，即使对同一个场景进行重建，高斯基元的分布也会比点云更加随机，这是由于所有的高斯基元只为渲染结果服务而忽视了几何结构与现实的一致性，对于渲染结果而言权重较高的高斯基元出现更频繁，因此保留这些重要高斯基元才能在GoF之间实现更高的匹配率。
\textbf{Sampling and multi-quality compression method}: To validate the effectiveness of our rendering-weight-aware sampling and compression scheme in preserving perceptually critical Gaussian primitives, comprehensive comparisons were conducted against multiple baseline sampling methods through quantitative evaluation of video quality and inter-frame mapping intensity (IFMI) \cite{livpformer}. Video quality assessment employs the PSNR between rendered 2D frames and corresponding ground truth training images under identical viewpoints. For IFMI analysis, temporal coherence across GoFs was quantified by measuring sequence consistency under varying distance and color variation value (DaCVV) thresholds. Evaluated baselines encompass uniform random sampling (URS) \cite{li2023viewport}, farthest point sampling (FPS) \cite{qi2017pointnet++}, local details preservation sampling (LPS) \cite{livpformer}, voxel sampling (VS) \cite{zhou2018voxelnet}, inverse density importance sampling (IDIS) \cite{groh2018flex}, geometric sampling (GS), and random sampling (RS). Figure \ref{sample1} demonstrates the PSNR degradation across different quality levels of tile compression, where each quality level corresponds to fixed compression ratios. Our method maintains superior rendering quality preservation during compression, particularly under aggressive compression ratios. This resilience stems from our selective retention of Gaussian primitives with higher rendering significance, effectively mitigating abrupt visual degradation when primitive counts fall below critical thresholds required for coherent scene representation. Figure \ref{sample2} illustrates the IFMI variations under different DaCVV thresholds across sampling methods. Our approach achieves substantially improved temporal coherence compared to alternatives, attributable to the inherent characteristics of 3DGS. Unlike point clouds that maintain geometrically consistent distributions, 3DGS primitives exhibit stochastic spatial arrangements optimized purely for rendering fidelity. High-weight Gaussians critical for view synthesis recur more persistently across frames, enabling our method to maintain superior GoF consistency through targeted primitive retention. This mechanism proves particularly crucial given 3DGS's rendering-oriented primitive distribution, where conventional geometry-preserving sampling strategies fail to capture perceptually vital elements.

%\textbf{Tiling method}: 为了验证我们的自适应切块方案面对3DGS视频流时的有效性，我们将我们的方法与\cite{li2022optimal}中的自适应方案（AT）进行了对比。该基线方法使用fast point feature histograms (FPFH)来判断不同切块的空间显著性，通过融合显著性来聚类对应切块。除此之外我们还比对了几种均匀切块方案，包括45T (the video is evenly divided into 45 tiles), 30T, 16T and NT (No Tiling)。我们从QoE结果与用户FoV Matching Rates两个方面去进行对比。所有的切块对比实验都是在Std5G的网络环境下进行的。如图\ref{Result2}所示，我们的方法在各类数据集上均取得了最好的QoE结果与FoV Matching Rates，这是因为过多的切块数量会消耗大量计算资源，而过少的切块数量会导致传输数据量的增大。相较于AT过于依赖高斯基元在空间上的分布来判断显著性的做法，我们的切块方案更能适应实际的用户FoV范围，在切块过程中更充分地考虑了视频特征。
\textbf{Tiling method}: To validate the efficacy of our adaptive tiling scheme for 3DGS video streaming, we compare our method with the adaptive tiling approach (AT) proposed in \cite{li2022optimal}. The baseline AT method employs fast point feature histograms (FPFH) to evaluate spatial saliency across tiles, clustering regions based on fused saliency scores. Additionally, we benchmark against uniform tiling configurations, including 45T (dividing the video into 45 uniform tiles), 30T (30 uniform tiles), 16T (16 uniform tiles), and NT (No Tiling, transmitting the entire video). Evaluations are conducted under Standard 5G (Std5G) network conditions, with QoE metrics and FoV matching rates as key performance indicators. As illustrated in Figure \ref{Result2}, our method achieves superior QoE and FoV alignment across all datasets. Excessive tiling granularity (e.g., 45T) introduces significant computational overhead, while insufficient partitioning (e.g., 16T or NT) increases redundant data transmission. Unlike AT, which predominantly relies on Gaussian spatial distributions for saliency estimation, our tiling scheme incorporates comprehensive video feature analysis during the partitioning process. This enables precise adaptation to real-world user FoV patterns, ensuring optimal resource allocation and perceptual quality.

%\textbf{Adaptive bitrate algorithm}: 为了验证我们基于元学习的ABR算法在泛用性以及小样本学习上的优越性，我们选用POT、QDA、HTS、AET四个点云视频的ABR算法作为基线对比，我们修改了这些算法中的部分设定使其能够适用于3DGS视频的传输。Figure \ref{ABR1}展示了在使用全部的训练数据（Std4G+Ext4G+Std5G+Ext5G）作为hybrid dataset的情况下，我们的模型在泛化到单个网络环境上能够取得最高的QoE表现，可以看到在网络环境差距巨大的情况下，我们的方法依然各个分环境下取得了最好的QoE结果，这是由于元学习即使在差异较大的数据集下依然可以总结跨环境的通用策略，模型不会过拟合到单个任务的噪声，从而能够从多样任务中提炼通用规律。为了测试来自单个网络环境的训练知识能否迁移到多个数据集，我们在Std5G的网络环境下训练模型，然后测试与训练模型在其他网络环境下的表现，结果如Figure \ref{ABR2}所示，虽然在差异较大的4G环境下预训练模型没有取得预期的效果，但仍然是所有方法里表现最好的，这验证了元学习在knowledge transfer上的有效性。Figure \ref{ABR3}和Figure \ref{ABR4}展示了在仅使用部分数据集训练的情况下模型的性能表现，可以看出相较于其他算法在数据集不足时出现的严重性能下滑，我们的方法在小样本上依然能取得很好的效果。以Std5G为例，当仅使用20%的训练数据时，在Std5G上就可以获得使用全部数据时84.9%的训练效果。而当使用50%的训练数据时，就已经相当于使用全部数据集94.3%的训练效果，远高于其他方法。
\textbf{Adaptive bitrate algorithm}: To validate the superior generalization capability and few-shot learning performance of our meta-learning-based ABR algorithm, we compare against four state-of-the-art point cloud video ABR baselines: POT\cite{li2023toward}, QDA\cite{yu2017qoe}, HST\cite{li2022optimal}, and RLA\cite{nguyen2023reinforcement}. These baselines were adapted with modified configurations to accommodate 3DGS video streaming. Figure \ref{ABR1} demonstrates that our model achieves the highest QoE performance when generalized to individual network environments using a hybrid training dataset (Std4G + Ext4G + Std5G + Ext5G). Our method maintains superior QoE across disparate network conditions, attributable to meta-learning’s ability to extract cross-environment universal policies while avoiding overfitting to task-specific noise. To evaluate cross-environment knowledge transferability, we trained the model exclusively on Std5G data and tested its performance under other network conditions (Figure \ref{ABR2}). While performance degrades in drastically different 4G environments, our method still outperforms all baselines, confirming meta-learning’s efficacy in knowledge distillation. Figures \ref{ABR3} and \ref{ABR4} illustrate performance under limited training data. Compared to severe performance degradation observed in baseline methods with insufficient data, our approach retains robust effectiveness in few-shot scenarios. For instance, using only 20\% of the Std5G training data achieves 84.9\% of the full-data performance, while 50\% training data reaches 94.3\% efficacy—significantly surpassing all comparative methods. This underscores our algorithm’s exceptional data efficiency and adaptability to resource-constrained scenarios.

\section{CONCLUSION}
%本文提出了一种全新的3DGS视频传输框架，探索了3DGS体积视频在实际应用中的潜力。在我们提出的框架中包含了我们的框架包含了从视频源处理到最优化传输至客户端的全部过程。我们首先提出了一个针对3DGS视频的显著性特征提取网络，基于时空显著性动态生成符合用户视觉习惯的自适应切块。然后针对现有动态3DGS方法依赖完整模型、无法兼容分块传输的问题提出了一种面向切块的动态3DGS编码方案，并且为每个切块构造了保证观看质量的多种质量等级版本。最后提出了一种基于元强化学习的自适应码率控制方案，我们设计了更适用于3DGS视频的QoE表达式，并通过元强化学习实现了在不同网络环境下的泛用性以及在小样本下的快速学习能力。The experimental results demonstrate the superiority of our proposal over existing schemes.
This study proposes a novel 3DGS video streaming framework that unlocks the practical potential of volumetric video transmission through three integrated innovations: a saliency-driven adaptive tiling mechanism employing spatiotemporal feature fusion and clustering algorithms to aggregate primitive blocks into viewport-optimized irregular tiles; a motion-categorized dynamic encoding scheme classifying tiles into static, low-dynamic, and high-dynamic types with corresponding shared or dedicated deformation fields, coupled with saliency-weighted multi-quality generation via adaptive Gaussian pruning; and a meta-reinforcement learning ABR controller incorporating 3DGS-specific QoE modeling for cross-environment generalization. Collectively, these components establish an end-to-end pipeline from source processing to optimized delivery, resolving critical barriers in practical 3DGS video deployment. The experimental results demonstrate the superiority of our proposal over existing schemes.

\ifCLASSOPTIONcaptionsoff
  \newpage
\fi

%\bibliography{reference}
\bibliographystyle{IEEEtran}


\end{document}